\documentclass[10pt,twocolumn,letterpaper]{article}

\usepackage[pagenumbers]{cvpr}

\definecolor{cvprblue}{rgb}{0.21,0.49,0.74}
\usepackage[pagebackref,breaklinks,colorlinks,allcolors=cvprblue]{hyperref}
\usepackage{multirow}

\def\paperID{5263} 
\def\confName{CVPR}
\def\confYear{2025}
\usepackage[accsupp]{axessibility}

\def\*#1{\mathbf{#1}}
\def\+#1{\mathcal{#1}}

\newcommand{\mrka}[1]{{\colorbox{red!30}{#1}}}
\newcommand{\mrkb}[1]{{\colorbox{red!20}{#1}}}
\newcommand{\mrkc}[1]{{\colorbox{red!10}{#1}}}

\title{LIRM: Large Inverse Rendering Model for Progressive Reconstruction of Shape, Materials and View-dependent Radiance Fields}

\author{
\!\!\!Zhengqin Li$^{1}$~
Dilin Wang$^{1}$~  
Ka Chen$^{1}$~ 
Zhaoyang Lv$^{1}$~ 
Thu Nguyen-Phuoc$^{1}$~ 
Milim Lee$^{1}$~ 
Jia-Bin Huang$^{12}$\\
Lei Xiao$^{1}$~ 
Cheng Zhang$^{1}$~ 
Yufeng Zhu$^{1}$~ 
Carl S. Marshall$^{1}$~ 
Yuheng Ren$^{1}$~ 
Richard Newcombe$^{1}$~ 
Zhao Dong$^{1}$ \\
$^{1}$Meta Reality Labs \qquad $^{2}$University of Maryland, College Park \\
}

\begin{document}
\twocolumn[{
     \renewcommand\twocolumn[1][]{#1}
     \maketitle
     \vspace{-12mm}
     \begin{center}
\includegraphics[width=1.0\linewidth]{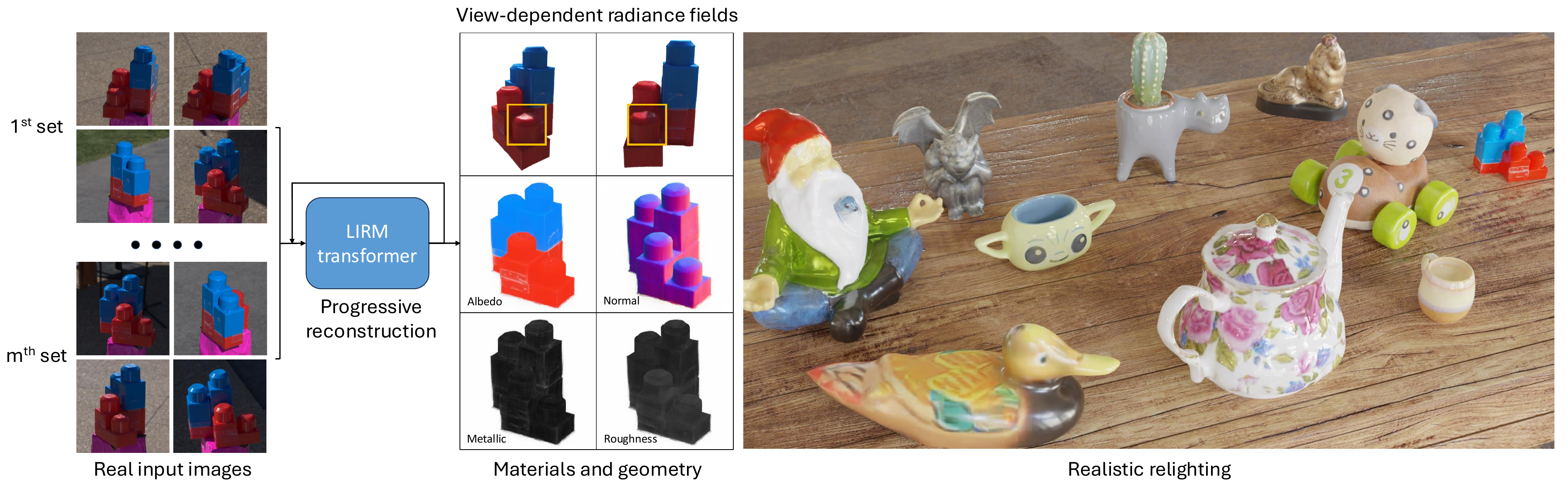}
\vspace{-5mm}
\captionof{figure}{
Given small sets of images (e.g., 4 to 8), LIRM progressively reconstructs view-dependent radiance fields, geometry and material reflectance in less than a second through a feed-forward transformer, enabling realistic rendering under novel lighting conditions. All relighting examples are reconstructed from real images in Stanford-ORB \cite{kuang2024stanford} and DTC datasets \cite{Dong_2025_CVPR}.}
\label{fig:teaser}
\end{center} 
     }]
\begin{abstract}
We present Large Inverse Rendering Model (LIRM), a transformer architecture that jointly reconstructs high-quality shape, materials, and radiance fields with view-dependent effects in less than a second. Our model builds upon the recent Large Reconstruction Models (LRMs) that achieve state-of-the-art sparse-view reconstruction quality. However, existing LRMs struggle to reconstruct unseen parts accurately and cannot recover glossy appearance or generate relightable 3D contents that can be consumed by standard Graphics engines. To address these limitations, we make three key technical contributions to build a more practical multi-view 3D reconstruction framework. First, we introduce an update model that allows us to progressively add more input views to improve our reconstruction. Second, we propose a hexa-plane neural SDF representation to better recover detailed textures, geometry and material parameters. Third, we develop a novel neural directional-embedding mechanism to handle view-dependent effects. Trained on a large-scale shape and material dataset with a tailored coarse-to-fine training scheme, our model achieves compelling results. It compares favorably to optimization-based dense-view inverse rendering methods in terms of geometry and relighting accuracy, while requiring only a fraction of the inference time.
\end{abstract}    
\section{Introduction}
\label{sec:intro}

\begin{figure*}
\centering 
\includegraphics[width=\textwidth]{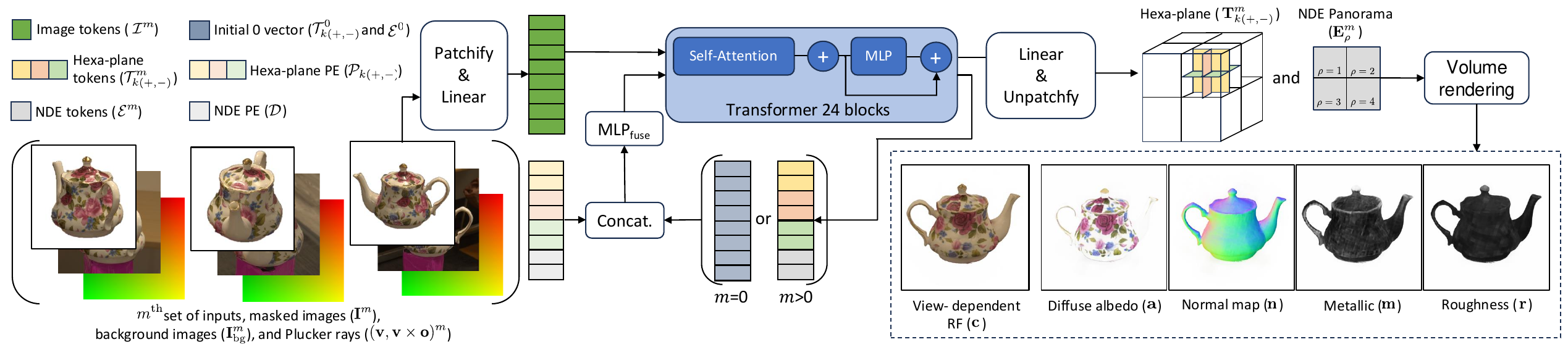}
\vspace{-0.3in}
\caption{The network architecture of LIRM. The inputs are masked images $\mathbf{I}^{m}$, background images to provide more lighting information $\mathbf{I}_{\text{bg}}^{m}$ and Pl\"ucker rays $(\mathbf{v}, \mathbf{v}\times \mathbf{o})^{m}$ that encodes camera intrinsics and extrinsics. These 3 images are concatenated together and turned into tokens through a simple linear layer. These tokens are sent to a self-attention transformer to update hexa-plane tokens ($\mathcal{T}_{k(+,-)}^{m}$, $k\in\{xy, xz, yz\}$) and NDE tokens ($\mathcal{E}^{m}$). We decode the 2 kinds of tokens into hexa-plane representation and NDE panoramas through linear layers, which can be used to render view dependent radiance fields and BRDF parameters through neural volume rendering. The decoded SDF volume can be used to extract accurate triangular mesh through standard marching cube.} 
\vspace{-0.1in}
\label{fig:network}
\end{figure*}

High-quality reconstruction of shape and materials from multi-view images, often referred to as inverse rendering, is a fundamental challenge in computer vision research. It has numerous important industrial applications in gaming, film, architecture, robotics, and AR/VR. Recent advancements in neural 3D representations \cite{lombardi2019neural,mildenhall2021nerf,muller2022instant,kerbl20233d} and generative modeling \cite{song2020score,ho2020denoising} have led to the development of more robust and efficient inverse rendering methods.
These advancements help democratize 3D content creation. 
Modern inverse rendering methods \cite{zhang2021nerfactor,zhang2021physg,zhang2022iron,boss2021nerd,boss2021neural,boss2022samurai,engelhardt2024shinobi,zhang2022modeling,jin2023tensoir,munkberg2022extracting,hasselgren2022shape,sun2023neural,liang2024gs,jiang2024gaussianshader,zhang2023nemf,yang2023sire,shi2023gir} can jointly reconstruct shape, materials and even lighting from multiple images captured in arbitrary natural illuminations with reasonable accuracy. This is often achieved by minimizing a rendering loss through a carefully designed optimization process. This is a great step forward from traditional measurement-based methods \cite{debevec2000acquiring,marschner1999image,matusik2003data} that require specialized devices and highly constrained environments. However, these optimization-based methods still suffer from long reconstruction time and require densely captured images. They also lack inductive bias to disambiguate lighting, materials and geometry, which often causes shadows and highlights to be baked into reconstructed materials. 

We present Large Inverse Rendering Model (LIRM), the first feed-forward transformer that takes less than 1 second to jointly reconstruct high-quality shape, materials, and view-dependent radiance fields of a full 3D object. It can achieve this from as few as 3 to 6 posed images captured in arbitrary, unconstrained environments. Without increasing GPU memory consumption, LIRM can progressively refine its prediction results by incorporating additional input views, allowing for the addition of missing textures and specular highlights in previously unseen regions or from novel view angles. Inverse rendering from sparse inputs is an extremely ill-posed problem. State-of-the-art optimization-based methods often fail to fully decompose materials, lighting, and geometry with dense observations. Our work draws inspiration from recent Large Reconstruction Models (LRM) \cite{hong2023lrm}, which are trained on large-scale 3D datasets \cite{deitke2023objaverse} and have achieved unprecedented high-quality sparse-view reconstruction results.

However, several drawbacks hinder their practical use in image-based 3D content creation frameworks. First, existing LRMs \cite{li2023instant3d,xu2023dmv3d,wang2023pf,wei2024meshlrm,zhang2025gs,xu2024instantmesh,openlrm,tochilkin2024triposr} output radiance fields without any view-dependent effects, which fails to correctly model glossy appearances. Secondly, they face difficulties in reconstructing unseen parts of objects. Unfortunately, naively adding more input views poses challenges due to GPU memory limitations and model capacity constraints during both training and inference. More importantly, they lack the ability to fully decompose geometry, materials, and lighting, which prevents them from generating relightable 3D contents that can be consumed by standard graphic pipelines.

In this work, we present a more practical model for efficient and robust reconstruction of high-quality relightable 3D contents. It fully supports downstream applications such as rendering, simulation and editing using standard graphics pipelines. Our approach introduces several important novel technical components. 
Firstly, we develop a novel network module that enables progressive updates of inverse rendering results. This is achieved by comparing predicted 3D tokens with new input image tokens using self-attention modules. As a result, our model can refine its predictions as more input views are added without increasing GPU memory consumption. 
Secondly, we propose a novel hexa-plane neural SDF representation. Our hexa-plane representation better reconstructs texture details and memorizes prior reconstruction results without significantly increasing computational cost.  Thirdly, we adopt neural directional encoding \cite{wu2024neural} into our feed-forward transformer architecture to recover view-dependent effects, which is essential for creating photorealistic appearances. Fourthly, we build a new large-scale 3D dataset with ground-truth materials and realistic appearances. This dataset carefully mimics real-world capturing settings to minimize domain gaps. We also made several improvements to further enhance reconstruction quality, including a higher-capacity model and an elaborate coarse-to-fine training paradigm that balances computational cost and quality. Experiments show that our model achieves competitive reconstruction results on real object benchmarks \cite{kuang2024stanford,ummenhofer2024objects}, and even outperforms several recent dense-view optimization-based methods in terms of geometry and relighting accuracy.

\section{Related Works}
\label{sec:related}

\paragraph{Inverse rendering} Built on recent advancements in neural 3D representations \cite{lombardi2019neural,mildenhall2021nerf,muller2022instant,kerbl20233d} and differentiable rendering \cite{li2018differentiable,zhang2019differential,KaolinLibrary,liu2019soft}, latest optimization-based invserse rendering methods \cite{zhang2021nerfactor,zhang2021physg,zhang2022iron,boss2021nerd,boss2021neural,boss2022samurai,engelhardt2024shinobi,zhang2022modeling,jin2023tensoir,munkberg2022extracting,hasselgren2022shape,sun2023neural,liang2024gs,jiang2024gaussianshader,zhang2023nemf,yang2023sire,shi2023gir} can reconstruct geometry, materials, and lighting from images densely captured in natural and unknown illumination. This is achieved by minimizing a rendering loss, a process takes from several minutes to hours. However, due to its ill-posed nature and the lack of effective inductive priors, even state-of-the-art inverse rendering methods are prone to generate artifacts under challenging scenarios, such as the presence of shadows, strong specular highlights, and interreflections. Various priors have been designed to improve inverse rendering accuracy. Earlier methods \cite{barron2013intrinsic,barron2014shape,lombardi2015reflectance} rely on hand-crafted regularizations. Recent learning-based methods increasingly rely on deep priors learned from large-scale real or synthetic datasets to solve challenging inverse rendering problems. These include single image intrinsic decomposition \cite{careaga2023intrinsic,meka2018lime,li2018cgintrinsics}, SVBRDF estimation \cite{li2018learning,li2018materials,li2017modeling,deschaintre2018single}, lighting estimation \cite{wang2021learning,li2020inverse,li2020openrooms,srinivasan2020lighthouse,gardner2017learning,garon2019fast}, and relighting \cite{xu2018deep,li2022physically}. However, none of the above methods can reconstruct fully relightable 3D objects from arbitrarily posed sparse inputs in a feed-forward manner.

\vspace{-0.2in}
\paragraph{Sparse-view reconstruction} Numerous attempts have been made to inject regularization \cite{jain2021putting,shi2024zerorf,niemeyer2022regnerf,kim2022infonerf} or learned priors \cite{chen2021mvsnerf,long2022sparseneus,johari2022geonerf,wang2021ibrnet,yu2021pixelnerf} into sparse-view neural radiance field reconstruction. However, these methods generally rely on sophisticated optimization paradigms with various regularization terms, which can take several minutes or even hours to reconstruct a 3D object with quality much lower than dense-view reconstruction methods.

Recent research has facilitated the use of diffusion priors \cite{song2020score,ho2020denoising} learned from billions of 2D images to significantly improve sparse-view reconstruction. These priors are usually applied either through score distillation sampling, which is usually time-consuming and unstable, \cite{poole2022dreamfusion,lin2023magic3d,wang2024prolificdreamer,wang2023steindreamer,wang2024taming,earle2024dreamcraft,wu2023hyperdreamer} or by fine-tuning pre-trained diffusion models on 3D object datasets to generate multi-view consistent images directly \cite{liu2023zero,long2024wonder3d,shi2023mvdream,liu2023syncdreamer,xu2024sparp,liu2024one,sargent2023zeronvs,tang2025mvdiffusion++}. While these methods demonstrate an impressive ability to hallucinate visually appealing appearances, the reconstructed unseen parts do not always align with real objects. They also lack the knowledge to faithfully recover material reflectance. In contrast, LIRM offers a more explicit approach to controlling the reconstruction process by adding additional input views. It achieves inverse rendering accuracy comparable to state-of-the-art methods optimization-baesd, making it a highly effective solution.

\vspace{-0.2in}
\paragraph{Large reconstruction models} 
LRM \cite{hong2023lrm} and its variants \cite{li2023instant3d,xu2023dmv3d,wang2023pf,wei2024meshlrm,zhang2025gs,xu2024instantmesh,openlrm,tochilkin2024triposr,tang2025mvdiffusion,sf3d2024} demonstrate transformer architecture's exceptional capability for sparse reconstruction. Trained on large-scale 3D datasets \cite{deitke2023objaverse,deitke2024objaverse} and multi-view image datasets \cite{yu2023mvimgnet}, they can reconstruct realistic geometry and texture details from very few or even a single image in a feed-forward manner within a second.

Most LRM variants focus on reconstructing radiance fields, which are incompatible with standard graphics pipeline for editing and visualization. While MeshLRM \cite{wei2024meshlrm} and InstantMesh \cite{xu2024instantmesh} extract mesh and texture maps, it remains a challenging problem to reconstruct relightable 3D objects with realistic specular highlights. Two concurrent works aim to solve this problem. RelitLRM \cite{zhang2024relitlrm} repurposes the LRM transformer as a neural renderer to directly predict view-dependent appearance under a new lighting condition. However, the output 3D Gaussian points have limitations, including a lack of support for near-field effects like area lighting and interreflections, as well as material and geometry editing applications. Similar to our method, SF3D \cite{sf3d2024} and AssetGen\cite{siddiqui2024meta} target fully decomposing materials, geometry and lighting. Compared to concurrent works, LIRM features a simpler network design with larger capacity, supporting multi-view progressive reconstruction and achieving lower relighting errors on a popular benchmark \cite{kuang2024stanford}.
\section{Method}
\label{sec:method}

Our LIRM network architecture is demonstrated in Fig. \ref{fig:network}. We start by introducing notations and preliminary knowledge of LRM. Then, we present our  update module for progressive reconstruction and an improved neural SDF representation for more detailed shape and material reconstruction. Next, we explain how we handle view-dependent radiance fields. Finally, we summarize our tailored coarse-to-fine training scheme and all the implementation details. 

\begin{figure}[t]
\centering
\includegraphics[width=\columnwidth]{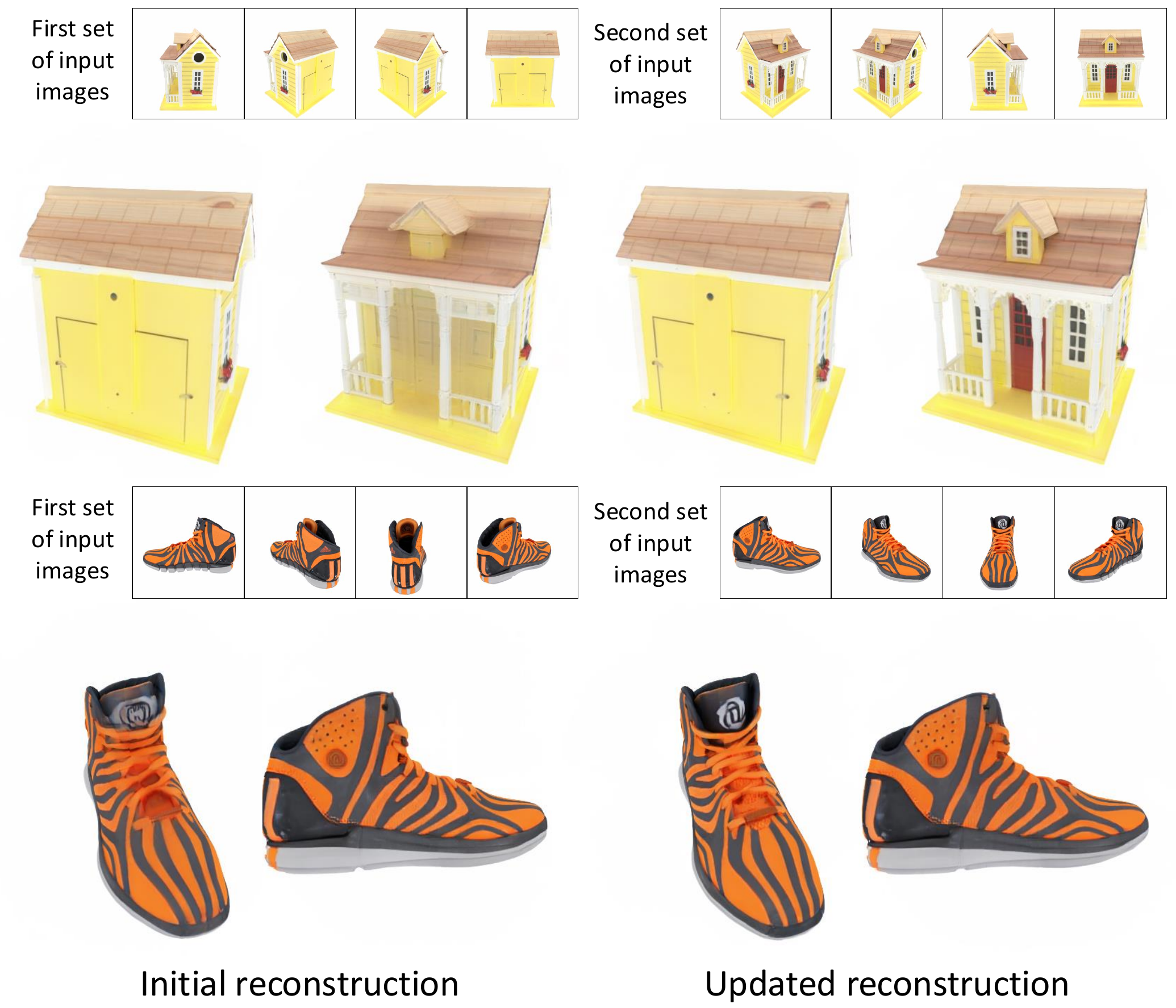}
\vspace{-0.3in}
\caption{Visualization of the initial and updated reconstruction. Our simple update strategy enables our model to memorize prior reconstruction while progressively improve results.}
\vspace{-0.1in}
\label{fig:update}
\end{figure}

\subsection{Preliminaries}
LIRM transformer architecture is built on MeshLRM \cite{wei2024meshlrm} but with larger capacity. Compared to the original LRM method \cite{hong2023lrm}, MeshLRM makes two major improvements in its network design. First, it uses Pl\"{u}ckers-rays representation instead of adaLN \cite{peebles2023scalable}. This enables better generalization to diverse camera settings and allows for cropping images to focus on the object, which is crucial for real-world applications with challenging camera settings (see Sec. \ref{sec:experiments}). Second, it removes the DinoViT \cite{caron2021emerging} encoder, which is difficult to train and results in reduced texture detail, likely due to its pre-training on semantic tasks. Instead, it uses a simple linear layer to tokenize image patches. More specifically, let $\{\*{I}\}$ be a set of non-overlapping $8\times 8$ image patches from multiple input views. Let $\{(\*v, \*v\times \*o)\}$ be the corresponding Pl\"ucker representation, where $\*v$ is the ray direction and $\*o$ is the camera original point. The image tokens are computed as 
\begin{equation}
\{\+I\} = \text{Linear}(\{\*I, (\*v, \*v\times \*o)\}).
\label{eq:lrm_im}
\end{equation}
MeshLRM adopts tri-plane as its 3D representation. Tri-plane is first tokenized with learned positional encoding $\{\+P_{k}\}, k\in\{xy, xz, yz\}$. These tri-plane tokens together with image tokens $\{\+I\}$ are sent to a simple transformer architecture that consists of a series of pre-LN self-attention blocks. Both types of tokens are updated by every self-attention block but we only keep the final output tri-plane tokens,
\begin{equation}
\!\!\!\!\!\!\{\+T_k\}\!=\!\text{Transformer}(\{\+P_{k}\},\{\+I\}).
\label{eq:lrm_vanilla}
\end{equation}
The predicted tri-plane tokens are then decoded by a simple linear layer. Each token is decoded to an $8\times 8$ non-overlapping feature patch on the tri-plane, 
\begin{equation}
\{\*T_{k}\} = \text{Linear}(\{\+T_{k}\}).
\label{eq:lrm_decode}
\end{equation}
The decoded $\{\*T_{k}\}, k\in\{xy, xz, yz\}$ is a standard tri-plane 3D representation, which can be used to render images either through volume ray marching or differentiable rasterization. In MeshLRM \cite{wei2024meshlrm}, the whole transformer is primarily trained with a rendering loss between rendered and ground-truth images, with several regularization terms to enforce geometry quality and training stability.

\begin{figure}
\centering
\includegraphics[width=\columnwidth]{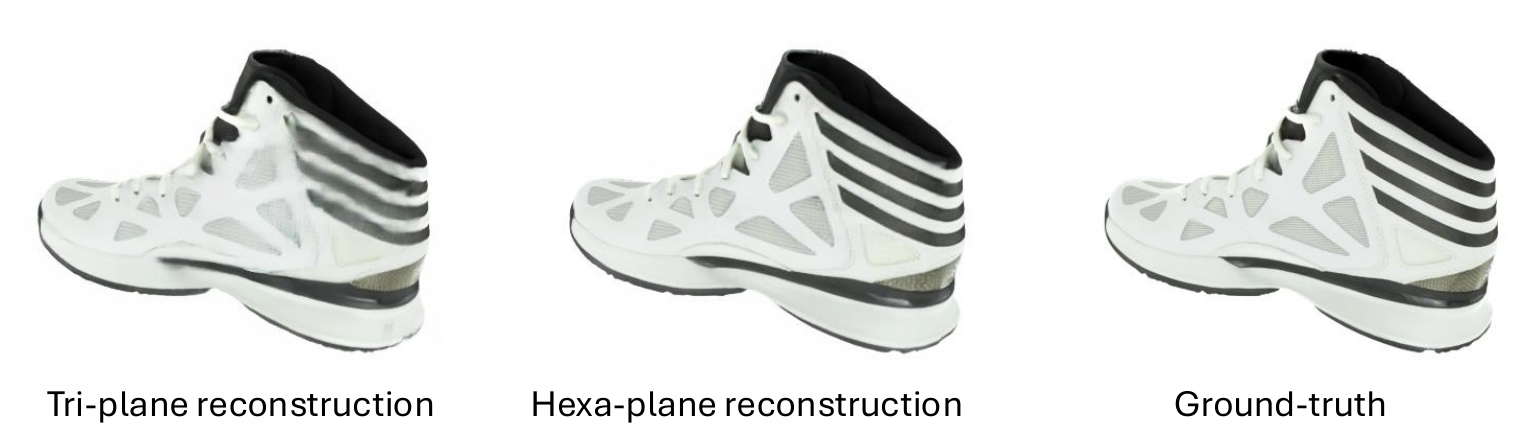}
\vspace{-0.3in}
\caption{Comparisons between our tri-plane and hexa-plane reconstruction. Here we show diffuse texture reconstruction results without considering material reflection and view-dependent effects. Hexa-plane clearly recovers better texture details.}
\vspace{-0.1in}
\label{fig:tri_vs_hexa}
\end{figure}
\

\subsection{Update Module for Progressive Reconstruction}
One unsolved challenge for existing LRMs is to reconstruct unseen parts of objects. Several attempts have been made to combine LRMs with multi-view diffusion models to hallucinate unseen appearance \cite{li2023instant3d,xu2024instantmesh}. However, the hallucinated appearance can not be guaranteed to align with the real object. Therefore, we argue that a more explicit way to control the reconstruction process is to enable users to interactively select more input views according to the current reconstruction result. Naively adding more images as inputs to the transformer will significantly increase the number of tokens, causing at least a linear increase of GPU memory consumption and close to quadratic increase of computation time, even with an advanced attention module \cite{dao2023flashattention}. On the contrary, our LIRM aims to support progressively adding an arbitrary number of images without increasing GPU memory consumption during both training and inference. 

The key to achieve our goal is to feed the predicted tri-plane tokens $\{\+T_{k}\}$ back into the transformer and update these tokens through self-attention with new input image tokens $\{\+I\}$. In this process, the learnable positional encoding $\{\+P_k\}$ are kept unchanged. We denote $\{\+T_{k}^{m-1}\}$ as the $(m-1)^{\text{th}}$ set of output tri-plane tokens predicted by the transformer. We compute the $m^{\text{th}}$ input tri-plane tokens $\{\+P_{k}^{m}\}$ by concatenating the $(m-1)^{\text{th}}$ output tokens with the learned positional encoding $\{\+P_{k}\}$ and pass it through a two-layer MLP. Eq. \eqref{eq:lrm_im} and \eqref{eq:lrm_vanilla} are re-written as:
\begin{eqnarray}
\{\+I^{m}\} &=& \text{Linear}(\{\*I^{m}, \*I_{\text{bg}}^{m}, (\*v, \*v\times \*o)^{m}\}), \\
\{\+P_{k}^{m}\} &=& \text{MLP}_{\text{fuse}}(\text{Concat}(\{\+T_{k}^{m-1}\}, \{\+P_{k}\})), \label{eq:lrm_fuse} \\ 
\{\+T_{k}^{m}\} &=&\text{Transformer}(\{\+P_{k}^{m}\},\{\+I^{m}\}) \label{eq:lrm_update},
\end{eqnarray}
where $\{\+T_{k}^{0}\} = \{\*0\}$. Since LIRM targets decomposing materials and lighting, we add images with background $\*I_\text{bg}$ as an extra input to help the network figure out the lighting condition of surrounding environments. We observe that this simple modification is sufficient to enable us to effectively update the tri-plane prediction with new input images without "forgetting" the prior observations. Fig. \ref{fig:update} shows an example of our reconstruction results where the first set of 4 input images only cover the front side of the object and the second set of 4 images only cover the back side. With the first set of 4 input images, our LIRM only reconstructs the front side of the object accurately. After taking the second set of inputs, our network updates the tri-plane prediction to obtain high-quality reconstruction of the full 3D object.

\subsection{Hexa-plane for Detailed Shape and Materials}

While tri-plane-based 3D representation can achieve highly detailed appearance and geometry in most scenarios, we observe that it struggles when both sides of an object contain complex but different textures, as shown in Fig. \ref{fig:tri_vs_hexa}. This limitation arises from using a single feature plane to represent both sides of textures in tri-plane representations. Therefore, we adopt a hexa-plane representation where we use 6 planes to divide the bounding box into 8 volumes, each with its own tri-plane. This representation utilizes the prior that the target object is likely to be roughly convex and located in the center of the 3D volume. K-plane \cite{fridovich2023k} also uses multiple planes to represent 3D objects and scenes, but its primary goal is to model the temporal axis for dynamic scene reconstruction.

We now present our neural SDF representation based on hexa-plane for joint shape and materials reconstruction. We denote our hexa-plane as  $\{\*T_{k(+,-)}\}$. The corresponding output tokens and positional encoding are defined as $\{\+T_{k(+,-)}\}$ and $\{\+P_{k(+,-)}\}$, which can be replaced into Eq. \eqref{eq:lrm_decode}, \eqref{eq:lrm_fuse} and \eqref{eq:lrm_update}. To render images and material maps from our hexa-plane representation, we use the SDF-based volume ray marching method proposed in \cite{yariv2021volume} for its simplicity and its ability to obtain high-quality geometry. We also considered \cite{wang2021neus} but it causes much higher computational cost, which will be discussed in the supplementary material. MeshLRM \cite{wei2024meshlrm} and InstantMesh \cite{xu2024instantmesh} use differentiable marching cube to reconstruct geometry and textures. However, that requires pre-training and special regularization terms to overcome training stability issues. Our training pipeline is much simpler while still showing promising geometry reconstruction and view synthesis results, as will be discussed in Sec. \ref{sec:experiments}. Formally, let $\*x=(x, y, z)$ be a 3D point. We can query feature $\*f_{xy}(\*x)$ from plane $\{\*T_{xy(+, -)}\}$ following, 
\begin{eqnarray}
\mathbf{f}_{xy}(\mathbf{x}) =  
\begin{cases}
\text{Bilinear}(\*T_{xy+}; x, y) & z \geq 0 \\
\text{Bilinear}(\*T_{xy-}; x, y) & z < 0 .
\end{cases}
\label{eq:hexa-query}
\end{eqnarray}
$\*f_{xz}$ and $\*f_{yz}$ can be queried similarly. $\*f_{xy}$, $\*f_{xz}$ and $\*f_{yz}$ are concatenated as feature vector $\*f$, which is sent to small MLPs with 2 to 3 hidden layers and 32 hidden dimensions for decoding, following \cite{wei2024meshlrm}. We use separate MLPs to predict SDF value, RGB color ($\*c$), normal ($\*n$), albedo ($\*a$), metallic ($\*m$) and roughness ($\*r$), written as, 
\begin{eqnarray}
\*s &=& \text{MLP}_{\*s}(\*f) + \*s_{\text{bias}}(\*{x})\\
\sigma &=& \begin{cases}
\frac{1}{2}\exp(-\frac{\*s}{\beta})   & s \geq 0 \\
1 - \frac{1}{2}\exp(\frac{\*s}{\beta})  & s < 0 
\end{cases} \label{eq:beta} \\
\*z &=& \text{sigmoid}(\text{MLP}_{\*z}(\*f)),~~\*z \in \{\*a. \*c, \*r, \*m\} \\ 
\*n &=& \text{normalize}(\text{MLP}_{\*n}(\*f)),
\end{eqnarray}
where $\sigma$ is density and $\beta$ is the standard deviation that controls sharpness of the reconstructed surface. We choose to gradually decrease $\beta$ as will be discussed in Sec. \ref{sec:training}. Meanwhile, $\*s_\text{bias}$ is a prior that we find important to ensure fast convergence of the training loss. We set 
\begin{equation}
\*s_\text{bias}(\*x) = ||\*{x}|| - 0.1R,
\end{equation}
where $R$ is the radius of the bounding sphere.

Fig. \ref{fig:tri_vs_hexa} compares our hexa-plane and tri-plane reconstruction results. We reduced the number of tokens per-plane (from $64 \times 64$ to $48 \times 48$) so that both representations have similar computational cost. We can clearly see that hexa-plane recovers much better texture details, while tri-plane causes texture patterns to ``leak" from one side to another, indicating a capacity limitation of tri-plane.

\subsection{View-dependent Radiance Fields}

\begin{figure}
\centering
\includegraphics[width=\columnwidth]{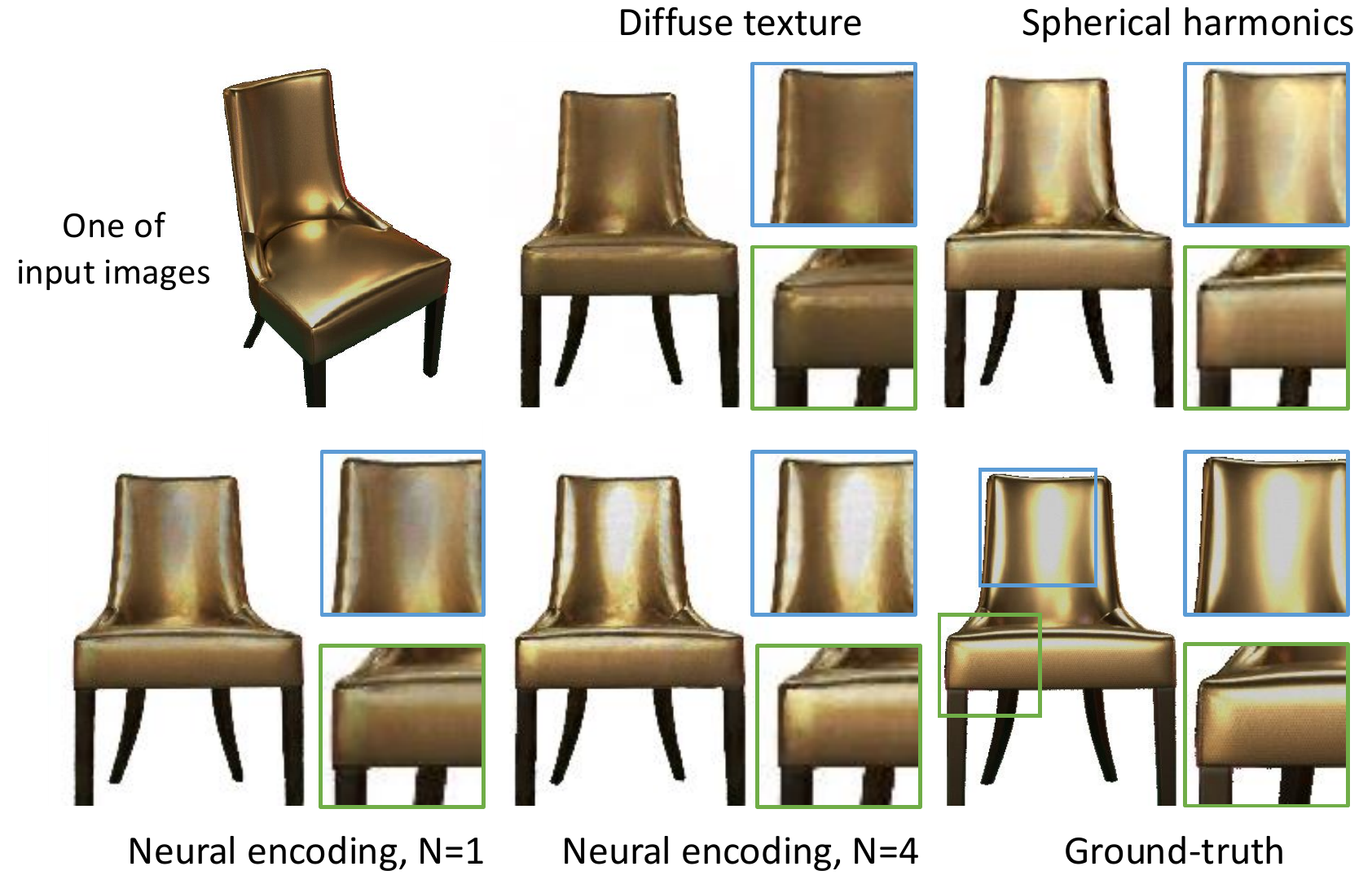}
\vspace{-0.3in}
\caption{Comparisons of different strategies to model view-dependent effect in a feed-forward network module.}
\vspace{-0.1in}
\label{fig:vd}
\end{figure}

View-dependent effect is important for modeling realistic appearance as glossy materials and specular highlights are commonly seen in daily objects. Moreover, neglecting view-dependent effects can negatively impact geometry quality, as the model will learn to create concave geometry to fake view-dependent appearance, as shown in \cite{zhang2025gs}. It is challenging to recover view-dependent radiance fields from sparse observations, especially when the 3D object is highly glossy. This requires a holistic understanding of not only geometry and materials but also lighting and the surrounding environment. Although the primary focus of LIRM is on reconstructing materials and geometry, we also investigate methods for reconstructing view-dependent radiance fields.

Simply adding the view direction as an extra input to the $\text{MLP}_{\*c}$ cannot work because of its limited capacity. We therefore explore two solutions. We first consider predicting three orders of spherical harmonic coefficients for every query point, following \cite{kerbl20233d} and \cite{yu2021plenoctrees}. However, this representation cannot handle high frequency angular signals, causing blurry specular highlights. To further improve the quality, we adopt the recent neural directional encoding (NDE) method \cite{wu2024neural} with modifications to make it compatible with a feed-forward transformer. NDE transfers the concept of feature volume into angular domain. It uses the reflection direction to query directional feature from a feature cubemap to model high frequency details in the angular domain.  However, this feature cubemap alone cannot model near-field reflections, which makes a large impact when the target object is not fully convex. \citet{wu2024neural} solve this problem by computing second bounce features, which is too expensive for training a feed-forward model. We solve the problem by predicting multiple NDE panoramas and let the model learn which panorama to query feature from. We predict those NDE panoramas progressively through our transformer architecture. Let $\{\+D\}$ be the learnable positional encoding for the NDE panorama and $\{\+E^{m-1}\}$ be the output NDE panorama tokens from $(m-1)^{\text{th}}$ update. We fuse them together similar to Eq. \eqref{eq:lrm_fuse} and send them to our self-attention transformer and decode the output tokens with a linear layer similar to  Eq. \eqref{eq:lrm_decode}: 
\begin{eqnarray}
\!\!\!\!\!\{\+D^{m}\} \!\!\!\!\! &=& \!\!\!\!\! \text{MLP}_{\text{fuse}}(\text{Concat}(\{\+E^{m-1}\}, \{\+D\})), \label{eq:lrm_fuse_d} \\ 
\!\!\!\!\!\{\+E^{m}\}, \{\+T_{k}^{m}\} \!\!\!\!\! &=& \!\!\!\!\! \text{Transformer}(\{\+P_{k}^{m}\}, \{\+D^{m}\}, \{\+I^{m}\}), \\
\{\*E_{\rho}^{m}\} \!\!\!\!\! &=& \!\!\!\!\! \text{Linear}(\{\+E^{m}\})
\end{eqnarray}
where $\rho \in [1, N]$ is the index of NDE panoramas and $N$ is the total number of NDE panoramas we use to approximate occlusion and near-field reflections. 

To utilize these NDE panoramas for rendering, we add a small MLP to predict NDE panorama index $\rho$. Let $\*v$ be the pixel ray direction and $\*l$ be the reflection direction. RGB color $\mathbf{c}$ at a 3D point $\*x$ can be written as:
\begin{eqnarray}
\*n &=& \text{normalize}(\text{MLP}_{\*n}(\*f)), \\
\*l &=& -2(\*v\cdot\*n)\*n + \*v \\
\phi, \theta &=& \text{arctan2}(\*l[0], \*l[1]), \text{arccos}(\*l[2]) \\
\rho &=& \text{sigmoid}(\text{MLP}_{\rho}(\*f)) \\
\*f_d &=& \text{Trilinear}(\{\*E_{\rho}^{m}\}_{\rho=1}^{N}; \rho, \theta, \phi) \\
\*c &=& \text{MLP}_{\*c}(\text{Concat}(\*f, \*f_d))
\end{eqnarray}

In Fig. \ref{fig:vd}, we compare the representation power of spherical harmonics and neural directional encoding in a feed-forward setting by overfitting LIRM to a single 3D object. For neural directional encoding experiments, we test $N=1$ and $N=4$. Spherical harmonics solution misses sharp specular highlights. NDE solution fails at concave regions when $N=1$ because we only model a single bounce. On the contrary, multiple panoramas ($N=4$) achieves the lowest reconstruction error and reconstructs accurate specular highlights for the whole object.

\subsection{Training Scheme}
\label{sec:training}

\paragraph{Coarse-to-fine training and acceleration} We adopt a coarse-to-fine training scheme to ensure fast convergence and reduce computational cost. In the first stage, we use a larger batch size and learning rate, but smaller resolutions and fewer samples per ray to obtain a coarse reconstruction result. In the second and third stages, we fine-tune the model by decreasing learning rates and batch sizes, while increasing resolutions and samples per ray to refine reconstruction details. We also gradually reduce the standard deviation $\beta$ by increasing $\frac{1}{\beta}$ following a linear schedule. Moreover, we increase the number of tokens for our tri-plane and hexa-plane representations compared to the prior state-of-the-art \cite{wei2024meshlrm}, which we rise from $32 \times 32 \times 3$ to $64 \times 64 \times 3$ and $48 \times 48 \times 6$ respectively. This makes training significantly slower at the third stage. Therefore, we incorporate occupancy grid acceleration from NerfAcc \cite{li2022nerfacc} into our differentiable renderer. Our final output plane resolutions are $512 \times 512$ and $384 \times 384$. For NDE panoramas, we add another $32\times 32$ tokens into our transfomer and decode them into 4 feature panoramas of resolution $128 \times 128$.  During training, we provide the update model with 2 sets of 3-6 images per iteration in the first stage, and 3 sets of images per iteration in the second and third stages. More details are in the supplementary material.

\vspace{-0.2in}
\paragraph{Loss functions} LIRM is trained with image losses computed from direct supervision of ground-truth 2D RGB image, normal and material maps. We use $L_2$ loss for all rendered 2D maps and add LPIPS \cite{zhang2018unreasonable} loss for RGB image and albedo, which we find essential to recover texture details. We also find that an $L_2$ loss on numerical normal computed from SDF gradients can help improve geometry accuracy. To compute the numerical normal, we perturb every 3D point in axis-aligned directions (i.e., along the x, y, and z axes) and then calculate the normalized gradients at each point, which serve as the numerical normal. We set the size of perturbation to be twice the size of a voxel in hexa-plane, following \cite{li2023neuralangelo}. However, this would require 3 times more feature query to compute numerical normal loss, which is expensive. We therefore only use it for the third stage of training, which we find to be sufficient. 

\vspace{-0.2in}
\paragraph{More implementation details} LIRM's transformer consists of 24 self-attention blocks, each with 16 heads and a feature dimension of 1024, where each head has a separate feature dimension of 64. Since LIRM's hexa-plane output both 2D images and material parameters maps, we increase the number of feature channels of $\{\*T_{k}\}$ from 32 to 64 compared to MeshLRM \cite{wei2024meshlrm}. We use AdamW optimizer with $(\beta_1, \beta_2) = (0.9, 0.95)$. The whole training takes around 2 weeks on 64 H100 GPUs. The inference time for one step is around 0.3 seconds on an A100 GPU.

\section{Experiments}
\label{sec:experiments}

\begin{table}
    \caption{Quantitative comparisons for view synthesis under uniform lighting on the \textbf{GSO} dataset. For the last two rows, we use the standard marching cube algorithm to extract triangular mesh from the predicted SDF volume.}
    \label{tab:quant_we_gso}
    \vspace{-0.1in}
    \setlength{\tabcolsep}{6pt}
    \renewcommand{\arraystretch}{0.4} 
    \small
    \centering
    \begin{tabular}{lccc}
    \toprule
    Radiance fields & PSNR ($\uparrow$) & SSIM ($\uparrow$) & LPIPS ($\downarrow$) \\
    \midrule
     MeshLRM \cite{wei2024meshlrm} & 28.13 & 0.923 & 0.093 \\
     GS-LRM \cite{zhang2025gs} & \mrkc{30.52} & \mrka{0.952} & \mrka{0.050} \\
     LIRM-hexa $1^{\text{st}}$  & 29.27 & 0.941 & 0.061  \\
     LIRM-hexa $2^{\text{nd}}$  & 30.48 & 0.947 & 0.056 \\
     LIRM-hexa $3^{\text{rd}}$  & \mrka{30.65} & \mrkb{0.949} & \mrkb{0.054} \\
     LIRM-hexa $4^{\text{th}}$  & \mrkb{30.56} & \mrkc{0.948} & \mrkb{0.054} \\
     LIRM-tri $4^{\text{th}}$ &  29.61 &  0.941 &  0.063 \\  
     \midrule
     Mesh & PSNR ($\uparrow$) & SSIM ($\uparrow$) &  LPIPS ($\downarrow$) \\
     \midrule
     MeshLRM \cite{wei2024meshlrm} & 27.93 & 0.925 & 0.081 \\
     LIRM-hexa $4^{\text{th}}$ & \mrka{29.22}  & \mrka{0.942} & \mrka{0.059} \\
     \bottomrule
    \end{tabular}
\end{table}

\begin{table}
    \vspace{-0.1in}
    \caption{Quantitative comparisons for view synthesis under uniform lighting on the \textbf{ABO} dataset. For the last row, we use the standard marching cube algorithm to extract triangular mesh from the predicted SDF volume.}
    \label{tab:quant_we_abo}

    \vspace{-0.1in}
    \setlength{\tabcolsep}{6pt}
    \renewcommand{\arraystretch}{0.4} 
    \small
    \centering
    \begin{tabular}{lccc}
    \toprule
        Radiance fields & PSNR ($\uparrow$) & SSIM ($\uparrow$) & LPIPS ($\downarrow$) \\
    \midrule
     MeshLRM \cite{wei2024meshlrm} & 28.31 & 0.906 & 0.108\\
     GS-LRM \cite{zhang2025gs} & 29.59 & 0.944 & \mrkb{0.051} \\
     LIRM-hexa $1^{\text{st}}$  & 32.69 & 0.957 & 0.056 \\
     LIRM-hexa $2^{\text{nd}}$  & \mrka{33.08} & \mrka{0.959} & \mrka{0.050} \\
     LIRM-hexa $3^{\text{rd}}$  & \mrkb{32.98} & \mrkb{0.958} & 0.055 \\
     LIRM-hexa $4^{\text{th}}$  & \mrkc{32.83} & \mrkb{0.958} & \mrkc{0.054} \\
     LIRM-tri $4^{\text{th}}$  & 32.58 & 0.956 & 0.055 \\ 
     \midrule 
     Mesh &  PSNR ($\uparrow$) & SSIM ($\uparrow$) & LPIPS \\
     \midrule
     LIRM-hexa $4^{\text{th}}$ & 29.79 & 0.951 & 0.069 \\
     \bottomrule
    \end{tabular}
\end{table}

\begin{table}
    \caption{Quantitative comparisons for view synthesis and inverse rendering under environment lighting on the \textbf{ABO} dataset.}
    \label{tab:quant_env_abo}
    \vspace{-0.1in}
    \setlength{\tabcolsep}{1.2pt}
    \renewcommand{\arraystretch}{0.4} 
    \small
    \centering
    \begin{tabular}{lccccccc}
    \toprule
     & \multicolumn{4}{|c|}{PSNR ($\uparrow$)} & \multicolumn{2}{|c|}{LPIPS ($\downarrow$)} & CD ($\downarrow$) \\
     \midrule
     LIRM & $\*c$ & $\*a$ & $\*r$ & $\*m$ &  $\*c$ & $\*a$ & - \\ 
     \midrule
     NDE $1^{\text{st}}$  & 29.01 & 32.34 & 23.20 & 27.28 & 0.064 & 0.070 & 0.123\\ 
     NDE $2^{\text{nd}}$  & \mrkc{29.70} & \mrkc{32.82} & \mrkc{23.30} & \mrkc{28.16} & \mrka{0.060} & \mrka{0.067} & 0.122\\
     NDE $3^{\text{rd}}$  & \mrkb{29.82} & \mrkb{32.85} & \mrka{23.36} & \mrkb{28.29} & \mrka{0.060} & \mrka{0.067} & \mrka{0.121}\\
     NDE $4^{\text{th}}$  & \mrka{29.92} & 32.76 & \mrkb{23.32} & \mrka{28.34} & \mrka{0.060} & \mrka{0.067} & \mrka{0.121}\\
     \midrule
     Diff $4^{\text{th}}$ & 29.50 & \mrka{32.95} & 23.05 & 28.14 & 0.061 & \mrka{0.067} & \mrka{0.121} \\  
     \bottomrule 
    \end{tabular}
\end{table}

\begin{table}
    \caption{Quantitative comparisons for view synthesis and inverse rendering under environment lighting on the \textbf{DTC} dataset.}
    \vspace{-0.1in}
    \setlength{\tabcolsep}{1.2pt}
    \renewcommand{\arraystretch}{0.4} 
    \small
    \centering
    \begin{tabular}{lccccccc}
    \toprule
     & \multicolumn{4}{|c|}{PSNR ($\uparrow$)} & \multicolumn{2}{|c|}{LPIPS ($\downarrow$)} & CD ($\downarrow$) \\
     \midrule
     LIRM & $\*c$ & $\*a$ & $\*r$ & $\*m$ &  $\*c$ & $\*a$ & - \\ 
     \midrule
     NDE $1^{\text{st}}$  & 27..59 & 30.19 & 18.71 & 26.97 & 0.092 & 0.096 & 0.119\\
     NDE $2^{\text{nd}}$  & 28.97 & 31.23 & \mrkc{18.79} & \mrkc{28.85} & 0.082 & 0.088 & 0.118\\
     NDE $3^{\text{rd}}$  & \mrkb{29.23} & \mrkc{31.50} & \mrka{18.88} & \mrkb{29.24} & \mrkb{0.081} & \mrkb{0.086} & \mrka{0.117}\\
     NDE $4^{\text{th}}$  & \mrka{29.35} & \mrkb{31.54} & \mrkb{18.84} & \mrka{29.33} & \mrka{0.080} & \mrka{0.085} & \mrka{0.117}\\
     \midrule
     Diff $4^{\text{th}}$ & \mrkc{28.98} & \mrka{31.60} & 18.44 & 29.12 & \mrkb{0.081} & \mrkb{0.086} & \mrka{0.117}\\  
     \bottomrule
    \end{tabular}
    \label{tab:quant_env_dtc}
\end{table}

\begin{figure}
\centering
\includegraphics[width=\columnwidth]{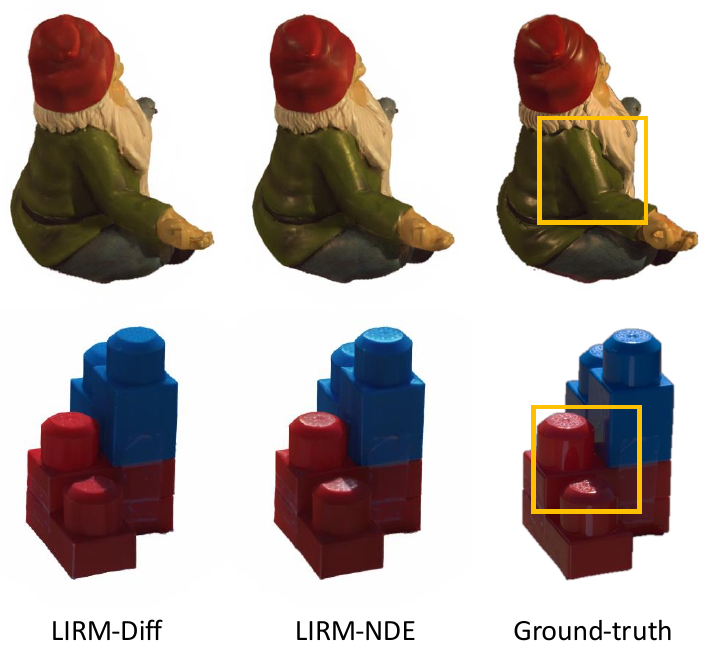}
\vspace{-0.3in}
\caption{LIRM-NDE can generalize to real images to recover view-dependent radiance fields.}
\vspace{-0.1in}
\label{fig:vd_real}
\end{figure}

\begin{table*}
\caption{Quantitative comparisons with methods on Stanford-ORB dataset \cite{kuang2024stanford} for relighting, view synthesis and geometry reconstruction. We separate methods into predictive and optimization-based methods. We select 3 top optimization-based methods from the leaderboard.}
\vspace{-0.1in}
\setlength{\tabcolsep}{3pt}
\renewcommand{\arraystretch}{0.6} 
\small
\centering 
\begin{tabular}{lccccccccc}
\toprule
\multirow{2}{*}{Methods} & \multicolumn{4}{c|}{Relighting} & \multicolumn{4}{c|}{View Synthesis} & Shape  \\
& PSNR-H ($\uparrow$) & PSNR-L ($\uparrow$) & SSIM ($\uparrow$) & LPIPS ($\downarrow$) &  PSNR-H ($\uparrow$) & PSNR-L ($\uparrow$) & SSIM ($\uparrow$) & LPIPS ($\downarrow$) & CD ($\downarrow$) \\ 
\midrule 
InvRender \cite{zhang2022modeling} & 23.76 & 30.83 & 0.970 & 0.046 & 25.91 & 34.01 & 0.977 & 0.042 & 0.44 \\ 
NVDiffrecMc \cite{hasselgren2022shape} & 24.43 & 31.60 & 0.972 & 0.036 & \mrkb{28.03} & \mrkb{36.40} & \mrkb{0.982} & 0.028 & 0.51 \\ 
Neural-PBIR \cite{sun2023neural} & \mrka{26.01} & \mrka{33.26} & \mrka{0.979} & \mrka{0.023} & \mrka{28.82} & \mrka{36.80} & \mrka{0.986} & \mrkb{0.019} & 0.43\\ 
\midrule 
MetaLRM \cite{siddiqui2024meta} & 21.46 & 28.00 & 0.956 & 0.045 & 19.93 & 26.20 & 0.956 & 0.042 & -\\ 
LIRM-diff $1^{\text{st}}$ & \mrkc{24.76} & \mrkc{32.11} & \mrkc{0.971} & 0.027 & 25.82 & 34.01 & 0.977 & 0.021 & 0.48 \\ 
LIRM-diff $3^{\text{rd}}$ & \mrkb{25.09} & \mrkb{32.45} & \mrkb{0.972} & \mrkb{0.025} & 26.66 & 34.88 & \mrkc{0.979} & \mrka{0.018} & \mrkc{0.38} \\ 
LIRM-NDE $1^{\text{st}}$ & 24.25 & 31.63 & 0.969 & 0.028 & 25.84 & 34.00 & 0.976 & 0.021 &  \mrkb{0.33} \\
LIRM-NDE $3^{\text{nd}}$ & 24.60 & 32.05 & \mrkc{0.971} & \mrkb{0.025} & \mrkc{27.03} & \mrkc{35.26} & \mrkc{0.979} & \mrka{0.018} & \mrka{0.31}  \\ 
\bottomrule
\end{tabular}
\label{tab:quant_env_orb}
\end{table*}

\begin{figure*}
\centering
\includegraphics[width=\textwidth]{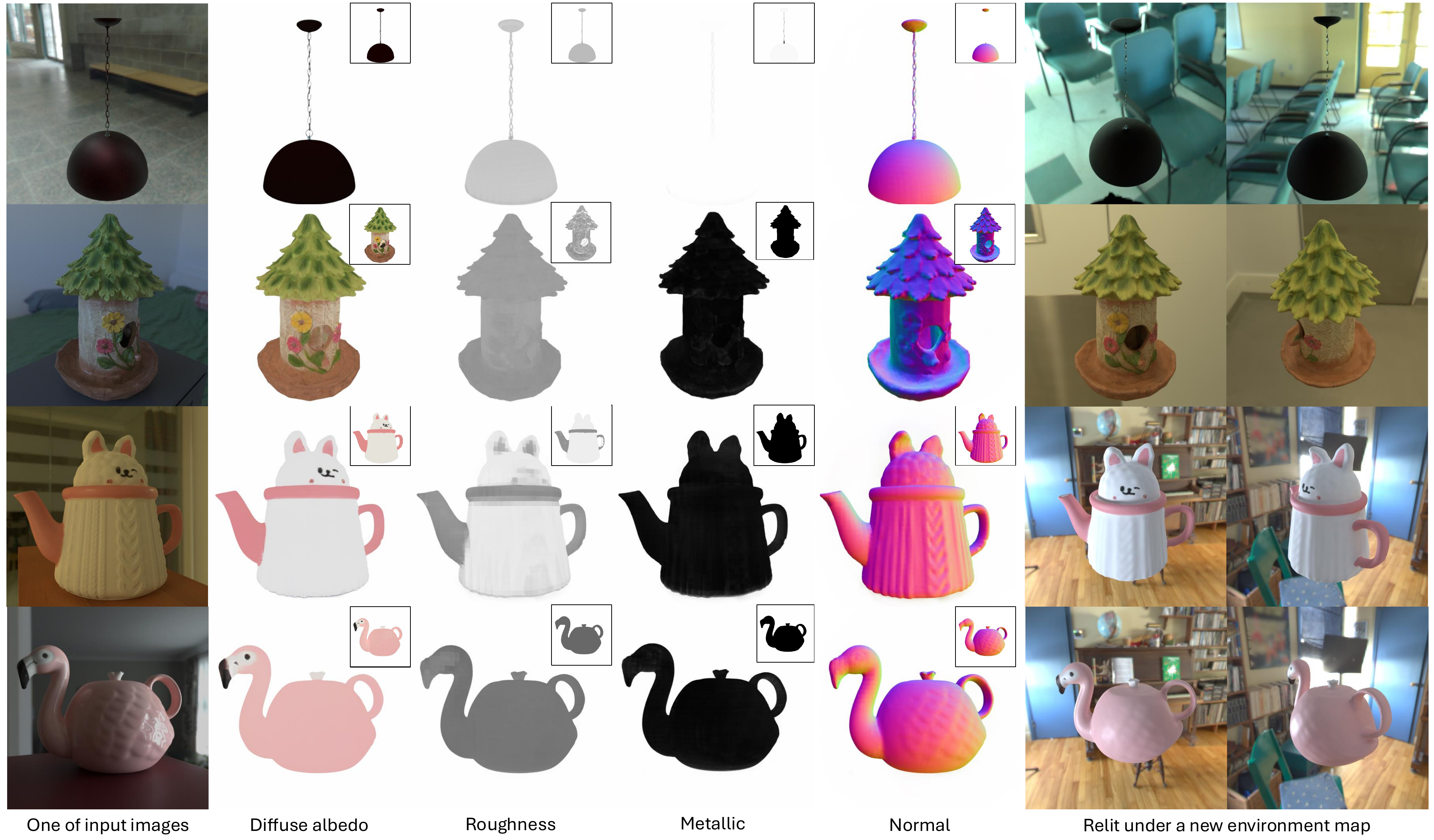}
\vspace{-0.3in}
\caption{Inverse rendering results on DTC \cite{Dong_2025_CVPR} (row 2 to 4) and ABO \cite{collins2022abo} (row 1) datasets. Material ground-truth are included in insets.}
\vspace{-0.1in}
\label{fig:exp_brdf_with_gt}
\end{figure*}

\begin{figure*}
\centering
\includegraphics[width=\textwidth]{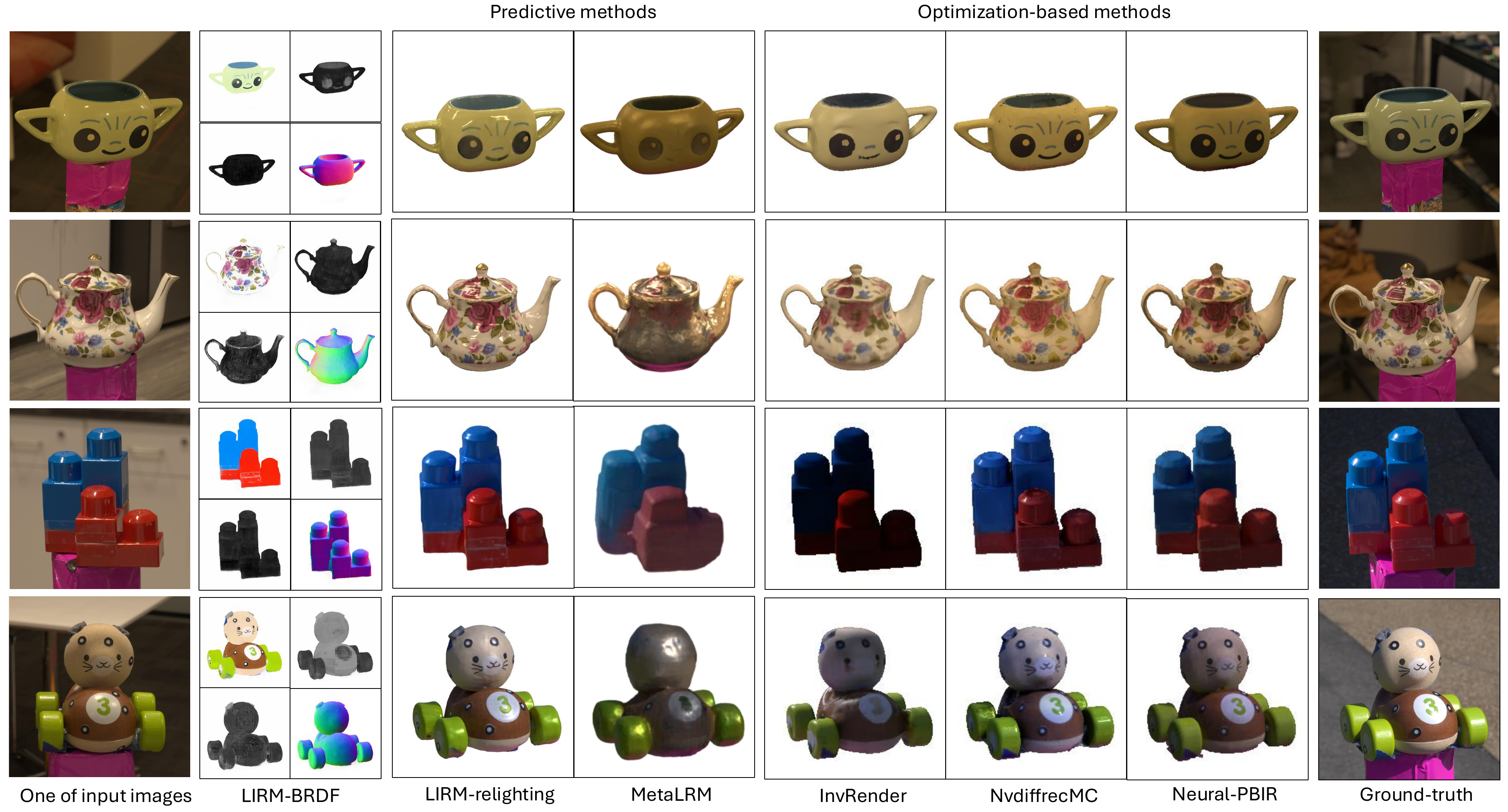}
\vspace{-0.3in}
\caption{Comparisons with prior works on the Stanford-ORB dataset \cite{kuang2024stanford}. }
\vspace{-0.1in}
\label{fig:stanford_orb}
\end{figure*}

\paragraph{Training data} We create a new dataset that carefully mimics real capturing environments to reduce domain gaps. We select 600k 3D objects from the Shutterstock dataset \cite{Shutterstock} with GT PBR materials. To render these 3D objects, we use 2.5K HDR environments collected from Laval Dataset and Polyhaven. Each HDR environment map is randomly rotated before rendering. We also randomly perturb the exposure time, white balance, camera intrinsics, and material parameters for data augmentation. Our final training set includes 38 million images with GT material maps. 

\vspace{-0.2in}
\paragraph{View synthesis under uniform lighting} We first train and test the LIRM model on datasets rendered with uniform lighting. This enables us to validate the effectiveness of our update module and hexa-plane representation. It also allows us to directly compare with state-of-the-art LRM variants. We tested on GSO and ABO datasets. Quantitative comparisons are summarized in Tab. \ref{tab:quant_we_gso} and Tab. \ref{tab:quant_we_abo}. Qualitative results can be seen in Fig. \ref{fig:update} and Fig. \ref{fig:tri_vs_hexa}. We tested 4 stages of update. The input and output views are selected following \cite{wei2024meshlrm}. We select 16 input views and 12 output views. For each stage, we randomly select 4 views from 16 input views. This is a more challenging setting compared to \cite{wei2024meshlrm} as it always uses canonical views as inputs. Nevertheless, both LIRM tri-plane and LIRM hexa-plane outperforms the prior stage-of-the-art volume-based LRM method \cite{wei2024meshlrm}. LIRM hexa-plane even outperforms the baseline with the first 4 input views, thanks to our larger number of plane tokens and the novel representation. We observe that our update model is the most effective within the first 3 sets of input views. LIRM hexa-plane achieves lower error compared to LIRM tri-plane but the gap is much smaller on ABO datasets, possibly because it has simpler textures.

\begin{figure}
\centering
\includegraphics[width=\columnwidth]{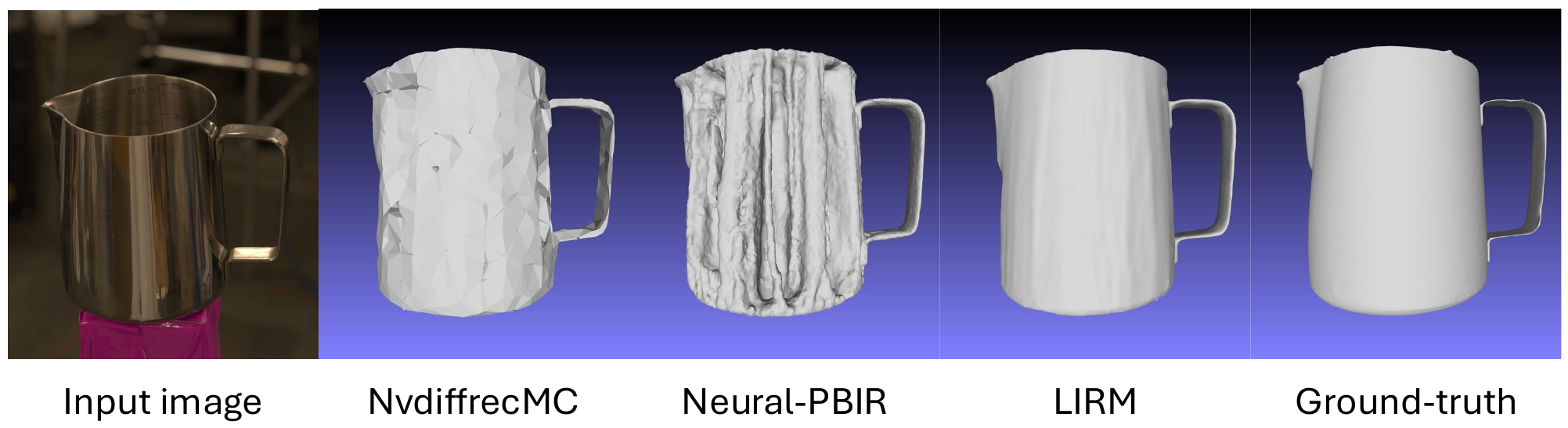}
\vspace{-0.3in}
\caption{Qualitative comparisons of geometry reconstruction quality. Meshes reconstructed by LIRM have much less artifacts compared to optimization-based methods.}
\vspace{-0.1in}
\label{fig:geometry}
\end{figure}

\vspace{-0.2in}
\paragraph{Inverse rendering and view synthesis under natural lighting} We first evaluate our LIRM on synthetic datasets. We randomly select 500 models each from ABO \cite{collins2022abo} and the newly released DTC \cite{Dong_2025_CVPR} datasets. Both datasets contain 3D models with high-quality material properties. We test two variants, LIRM-Diff only predict diffuse texture and material parameters. LIRM-NDE also predicts NDE panoramas to model view-dependent effects. The view selection is the same as previous experiments. Quantitative results are summarized in Tab. \ref{tab:quant_env_abo} and Tab. \ref{tab:quant_env_dtc}. From quantitative results, we observe that LIRM-NDE consistently achieve lower view synthesis errors compared to LIRM-Diff, indicating that our NDE module can successfully model view-dependent effects. Geometry and material quality of the 2 models are similar. Similarly, LIRM-Diff and LIRM-NDE can improve reconstruction quality with more inputs. Fig. \ref{fig:exp_brdf_with_gt} shows that our material prediction can accurately match the ground-truth, even for spatially varying roughness (the third row) and metallic objects (the first row). As we encode the background image $\*I_{\text{bg}}$ as one of LIRM's inputs to help it better understanding surrounding environments, our models works exceptionally well in separating lighting color from materials, leading to highly realistic object relighting results (the third row, last 2 columns). 

We then test both two variants on a real dataset, Stanford-ORB \cite{kuang2024stanford}. For each object in the Stanford-ORB dataset, we randomly select 18 images and evenly divide them into 3 sets as inputs to our LIRM models. Objects in the Stanford-ORB dataset usually only occupy a small central region of input images. Directly resizing the images will cause loss of details. We utilize the flexibility of Pl\"{u}cker-rays representation by first compute the Pl\"{u}cker rays for the whole image and then only crop and resize the region of interest based on the object's foreground mask as inputs to our models. This allows us better usage of network capacity and works well in practice. With this simple modification, LIRM generalizes impressively to real data. Quantitative comparisons on relighting, view synthesis, and geometry reconstruction accuracy are summarized in Tab. \ref{tab:quant_env_orb}. For all relighting results, we first run the standard marching cube algorithm to extract triangular mesh from our SDF volume and then extract BRDF texture maps by querying BRDF values for surface points, following \cite{sun2023neural}.  LIRM achieves reconstruction quality on par and even better than state-of-the-art optimization-based methods, which takes dense views as inputs and several hours to run. Even with one set of inputs, LIRM achieves the second best relighting quality. With 3 sets of inputs, it achieves the highest geometry reconstruction quality. Fig. \ref{fig:stanford_orb} shows LIRM can better handle specular materials compared to optimization-based methods. For example, in row 1 and 2, the specular highlights on the surface of our reconstructed cup and teapot are realistic and can closely match those of ground-truth images. While optimization-based methods either miss the specular highlights or cannot match the ground-truths accurately. We also compare to concurrent LRM-based inverse rendering method \cite{siddiqui2024meta}. LIRM outperforms by a large margin both qualitatively and quantitatively. Fig. \ref{fig:geometry} demonstrates that LIRM can achieve accurate geometry reconstruction even for highly specular objects, while state-of-the-art optimization-based inverse rendering methods \cite{sun2023neural,hasselgren2022shape} tend to generate artifacts. Fig. \ref{fig:vd_real} shows that with real inputs, LIRM can also effectively handle view-dependent effects. 

\begin{figure}
\centering
\includegraphics[width=\columnwidth]{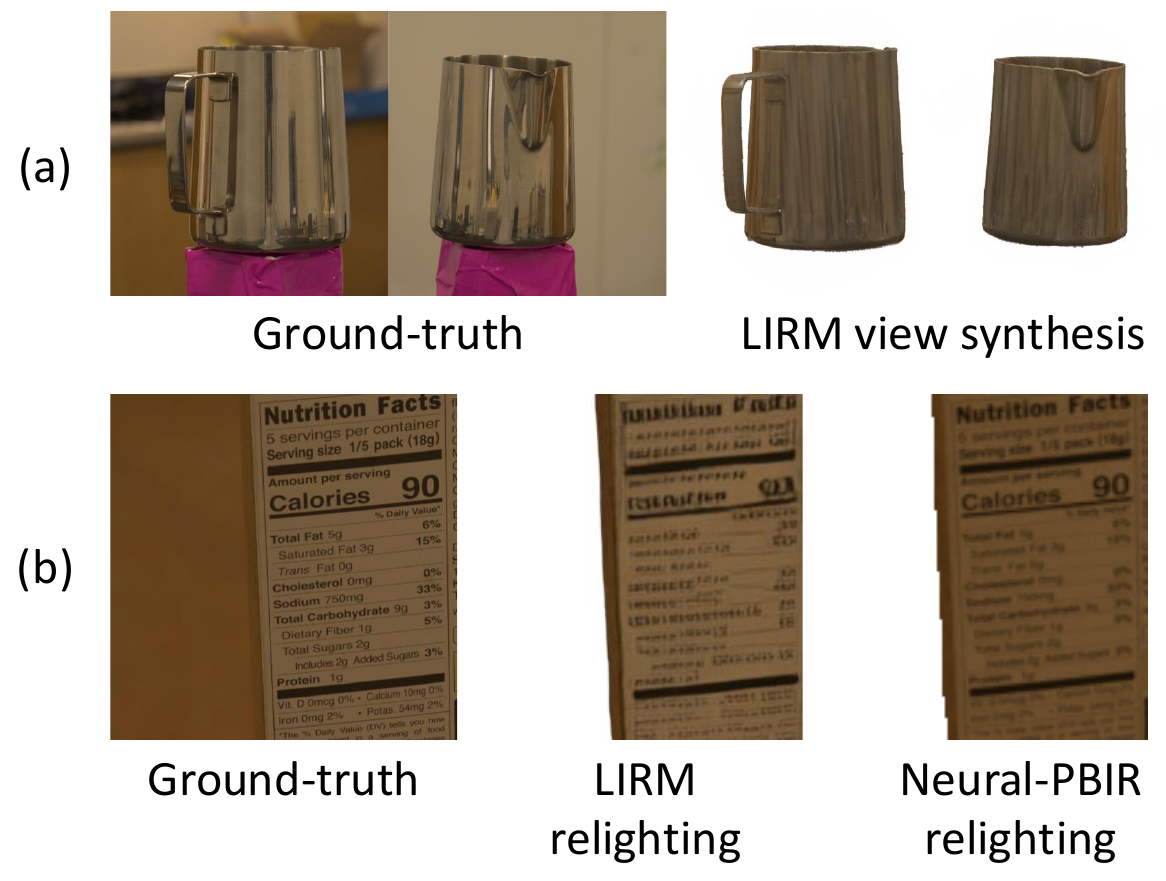}
\vspace{-0.3in}
\caption{Two limitations of LIRM. (a) LIRM-NDE cannot handle high-frequency reflection of mirror surfaces. (b) Compared to optimization-based method, LIRM still traces behind in reconstructing texture details.}
\label{fig:limitations}
\vspace{-0.1in}
\end{figure}

\vspace{-0.2in}
\paragraph{Limitations of LIRM} We observe two major limitations of LIRM. First, even through our NDE module can model specular highlights and view dependent effects, as shown in Fig. \ref{fig:vd_real}, Fig. \ref{fig:vd} and Tab. \ref{tab:quant_env_orb}, it fails to model mirror reflection as shown in Fig. \ref{fig:limitations} (a). We argue this is an extremely challenging problem as it requires the network to reconstruct the full 3D scene from sparse observation of reflection of an unknown object. In addition, compared to optimization-based methods, LIRM still fails to recover more detailed texture. In Fig. \ref{fig:limitations} (b), LIRM can reconstruct writings for the brand name printed on the box, but not ingredient list, unlike an optimization-based method \cite{sun2023neural}. We attribute this limitation to the network capacity, as our hexa-plane representation should have a sufficiently high resolution. A larger model may be required to achieve higher-quality reconstructions. 

\begin{figure}
    \centering
    \includegraphics[width=\columnwidth]{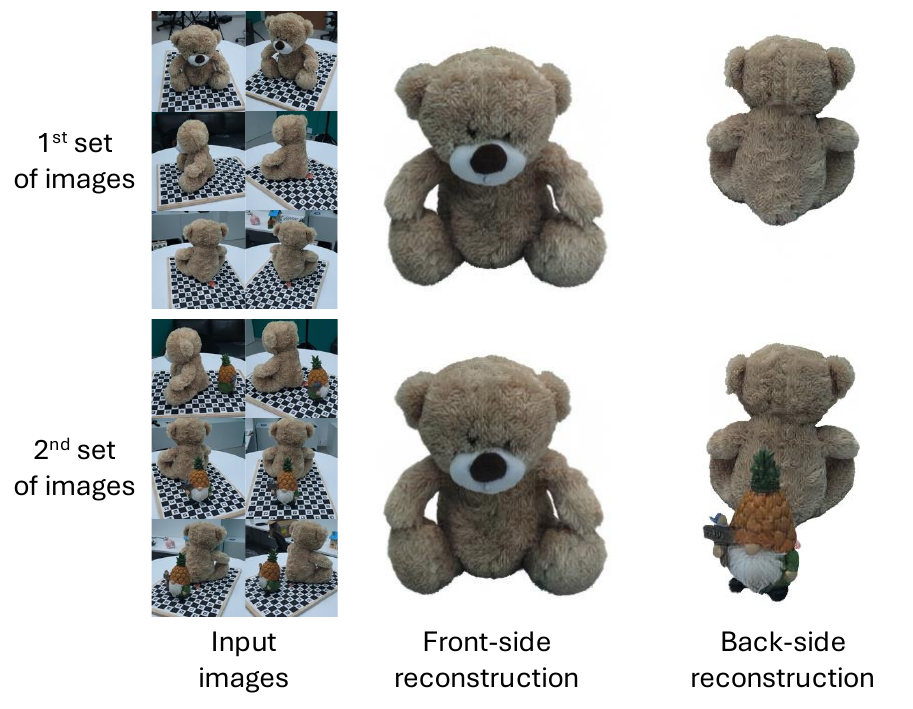}
    \vspace{-0.3in}
    \caption{Reconstructing a changing scene with the LIRM update model. Even if we change the scene configuration after capturing the first set of input images, LIRM's update model can still achieve accurate reconstruct of the newly added object while preserving old reconstruction.}
    \vspace{-0.1in}
    \label{fig:changing_scene}
\end{figure}

\begin{figure}
    \centering
    \includegraphics[width=\columnwidth]{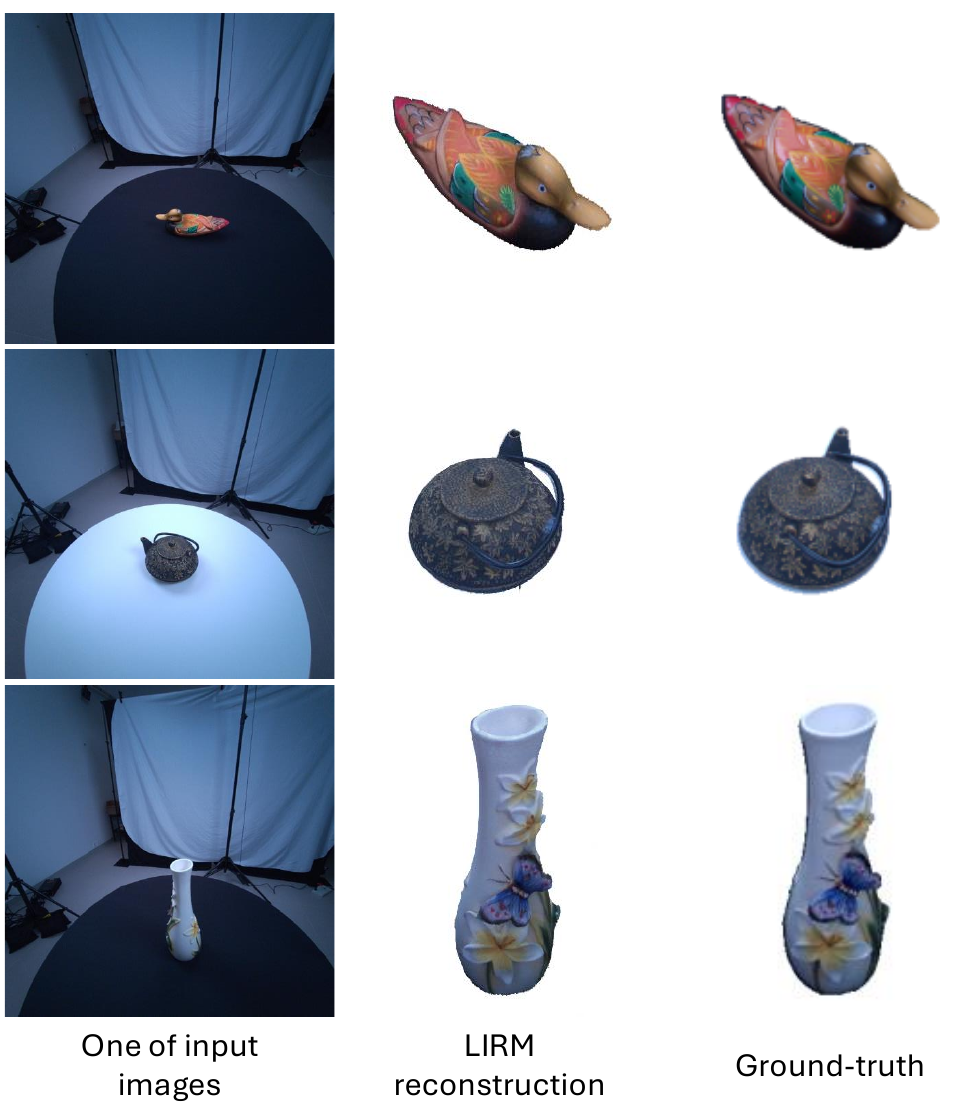}
    \vspace{-0.3in}
    \caption{LIRM reconstruction from images casually captured using egocentric Aria glasses \cite{engel2023aria}.}
    \vspace{-0.1in}
    \label{fig:aria}
\end{figure}

\vspace{-0.2in}
\paragraph{Challenging scenarios} We test our LIRM's generalization ability on two more challenging scenarios. The first scenario is updating reconstruction of a changing scene, where we first capture one set of images, then change the scene configuration by adding a new object and capture the second set of images. We use the two sets of images as inputs to LIRM's update model. The input images and reconstruction results are shown in Fig. \ref{fig:changing_scene}. Despite that this scenario never occurs in the training data, LIRM manages to reconstruct the added object accurately while still preserving the initial reconstruction of the first object. Note that the front face of the "teddy bear" is not shown in the second set of input images and yet our model keeps all the facial details unchanged through the updating process. This suggests that our model may be applied to dynamic scene reconstruction or large-scale scene reconstruction when one set of images cannot cover the whole scene. 

In the second scenario, we test our LIRM model on images casually captured by egocentric Aria glasses. Users were asked to wear a pair of Aria glasses, walking towards the object, causally look around the object and then walk away. This egocentric capturing setting better mimics how common people may take photos for 3D reconstruction. However, it also presents unique challenges, such as large field-of-view, motion blur, sensor noise, etc. We directly test our LIRM model on the challenging egocentric captured images without any fine-tuning. Example inputs and reconstruction results are shown in Fig. \ref{fig:aria}. For each video sequence, we extract 16 images as inputs. Even though there are clear domain gaps between testing inputs and our training data, our LIRM still reconstructs the object appearance that is very close to the ground-truth. 

\section{Conclusion}
We present LIRM, a Large Inverse Rendering Model that rapidly reconstructs high-quality shape, materials, and radiance fields with view-dependent effects from sparse inputs in under one second. LIRM overcomes the limitations of existing LRMs by introducing three key technical contributions: an update model for progressive reconstruction, a hexa-plane neural SDF representation for detailed texture recovery, and a novel neural directional encoding mechanism for view-dependent effects. Trained on a large-scale dataset in a coarse-to-fine manner, LIRM delivers results comparable to optimization-based methods, while significantly reducing inference time.

\vspace{-0.2in}
\paragraph{Supplementary} We will include more implementation details, qualitative and quantitative analysis on multi-stage training and camera trajectories.  We will add comparisons with a MeshLRM \cite{wei2024meshlrm} baseline and with optimization-based inverse rendering methods on BRDF reconstruction accuracy.

\vspace{-0.2in}
\paragraph{Acknowledgements} We thank Yawar Siddiqui, Jesus Zarzar and David Novotny for providing baseline comparisons with MetaLRM \cite{siddiqui2024meta}. We thank Yunzhi Zhang, Hong-Xing Yu, Guangyan Cai and Chen Sun for supporting experiments on the Stanford-ORB dataset \cite{kuang2024stanford}.
{
    \small
    \bibliographystyle{ieeenat_fullname}
    \bibliography{main}
}
\clearpage
\setcounter{page}{1}
\maketitlesupplementary

\section{Overview}
\label{sec:rationale}
Our supplementary material consists of three parts. 
\begin{itemize}
\item Implementation details, including multi-stage training, accelerated deferred rendering, training and testing datasets creation.
\item Ablation studies on synthetic dataset, including impacts of multi-stage coarse-to-fine training and input camera trajectories.
\item Quantitative comparisons with a MeshLRM \cite{wei2024meshlrm} baseline. 
\item Comparisons with optimization-based inverse rendering methods on BRDF reconstruction accuracy.  
\end{itemize}
All of the results in the supplementary materials use the same implementation and datasets as the main paper. 

In addition, we include a video for better visualization. 

\section{Implementation details}

\paragraph{Coarse-to-fine training}
Our training consists of three stages. We first train with large batch sizes but small resolutions for fast convergence, and later train with high resolutions but small batch sizes for better details. The hyper-parameters for the three stages are summarized in Tab. \ref{tab:coarse_to_fine_hype}. Similar to \cite{wei2024meshlrm}, we utilize cropped regions from the original ground-truth image for supervision. $\beta$ is the standard deviation that controls sharpness of the surface, as mentioned in Eq. \eqref{eq:beta}. We increase $\frac{1}{\beta}$ linearly following \cite{sun2023neural}. We also tried to learn $\beta$ using gradient descent; however, this approach resulted in less stable training.

\vspace{-0.2in}
\paragraph{Accelerated deferred rendering}
Deferred rendering \cite{zhang2022arf} is used in all prior volume-based LRM methods \cite{hong2023lrm,li2023instant3d,wei2024meshlrm} to reduce GPU memory consumption. The basic idea is to cache the gradients so that we can render an image patch-by-patch while still computing a perceptual loss like LPIPS on the whole image, which is essential for reconstructing texture details. In our setting, we find that rendering the whole image crop in a single pass can reduce time consumption for the first two stages of training, making deferred rendering unnecessary. However, in the third training stage, rendering the whole image crop becomes impractical due to memory overflow, as we significantly increase the number of samples per ray and compute numerical normals for improved geometry reconstruction. We therefore adopt the occupancy grid acceleration developed in Nerfacc \cite{li2022nerfacc} in the third stage. Before we render the image crops, we first compute an occupancy grid of resolution 250 and filter out voxels with $\alpha$ lower than $1e^{-4}$. The computation of occupancy grid takes only 0.05 s, while significantly reducing GPU memory consumption and accelerates training. It allows us to filter out 91.2 percentage of sampled points on average and reduce the time consumption to render four image crops of size $192 \times 192$ from around 4s to 0.3s. 

\vspace{-0.2in}
\paragraph{Training datasets creation} The camera settings used to create synthetic datasets under uniform lighting are identical to those used under environmental lighting. For each 3D model, we render 32 images for training. To ensure generalizability to various camera types, the field of view is uniformly sampled between $15^{\circ}$ and $85^{\circ}$. The elevation angle is uniformly sampled between $[-5^{\circ}, 70^{\circ}]$ while azimuth angle is uniformly sampled between $[0^{\circ}, 360^{\circ}]$. For data augmentation,  we employ an auto-exposure algorithm that automatically adjusts the camera's exposure settings. During training, we also on the fly apply a perturbation scale to image pixels uniformly sampled between 0.75 and 1.25. For the synthetic dataset under environmental lighting, we generate two versions: one with the original roughness values from the 3D models, and another where the roughness values are scaled by a factor between 0.3 and 0.6 to create more specular appearances. The two datasets are mixed together to train LIRM for inverse rendering. 

\begin{table*}[t]
\centering
\begin{tabular}{|c|c|c|c|c|c|c|c|c|c|c|c|}
\hline
LRM    & Learning & Batch & Input  & Samples &  Input & GT & Crop. & \multirow{2}{*}{$\frac{1}{\beta}$} & \multirow{2}{*}{Epochs} & \multirow{2}{*}{Update} \\ 
-VolSDF    & rate  & size & num.  & per ray & res. & res.& res.& & & \\
\hline
Stage 1 & $4e^{-4}\!\!\!\!\rightarrow\!2e^{-5}$ & 768 & [3, 6] & 128 & 512 & 256 & 128 & $1e\!\rightarrow\!2e^{2}$ & 30 & 2 \\
\hline
Stage 2  & $2e^{-5}\!\!\!\!\rightarrow\!1e^{-6}$ & 320 & [3, 6] & 512 & 512 & 384 & 128 & $2e^2\!\rightarrow\!2.5e^2$ & 5 & 3 \\
\hline
Stage 3 & $1e^{-6}\!\!\!\!\rightarrow\!0$ & 256 & [3, 6] & 1024 & 512 & 512 & 192 & $2.5e^{2}$ & 2 & 3 \\ 
\hline
\end{tabular}
\vspace{-0.1in}
\caption{Training settings for LIRM. Our learning rate decreases following a cosine scheduling while $\frac{1}{\beta}$ increases following a linear scheduling.}
\vspace{-0.1in}
\label{tab:coarse_to_fine_hype}
\end{table*}

\vspace{-0.2in}
\paragraph{Testing datasets creation} We select input and output views for the testing datasets in a manner that closely follows MeshLRM \cite{wei2024meshlrm}. We set elevation angles as $0^{\circ}$, $20^{\circ}$, $40^{\circ}$ and uniformly divide azimuth angle into 16 intervals, which gives us 48 views in total. From these 48 views, we uniformly sample 8 views at elevation $20^{\circ}$ and $40^{\circ}$ as the input views. We then sample 12 views from the remaining 32 views as the output views. The FoV is set to be $50^{\circ}$. The camera always looks at the object center. Its distance to the object center is set as the minimal distance that can cover the object's bounding sphere. esting datasets captured under uniform lighting and environmental lighting conditions follow the same camera settings and view selection method.

\section{Experiments on Synthetic Data}

\begin{figure}
\centering
\includegraphics[width=\columnwidth]{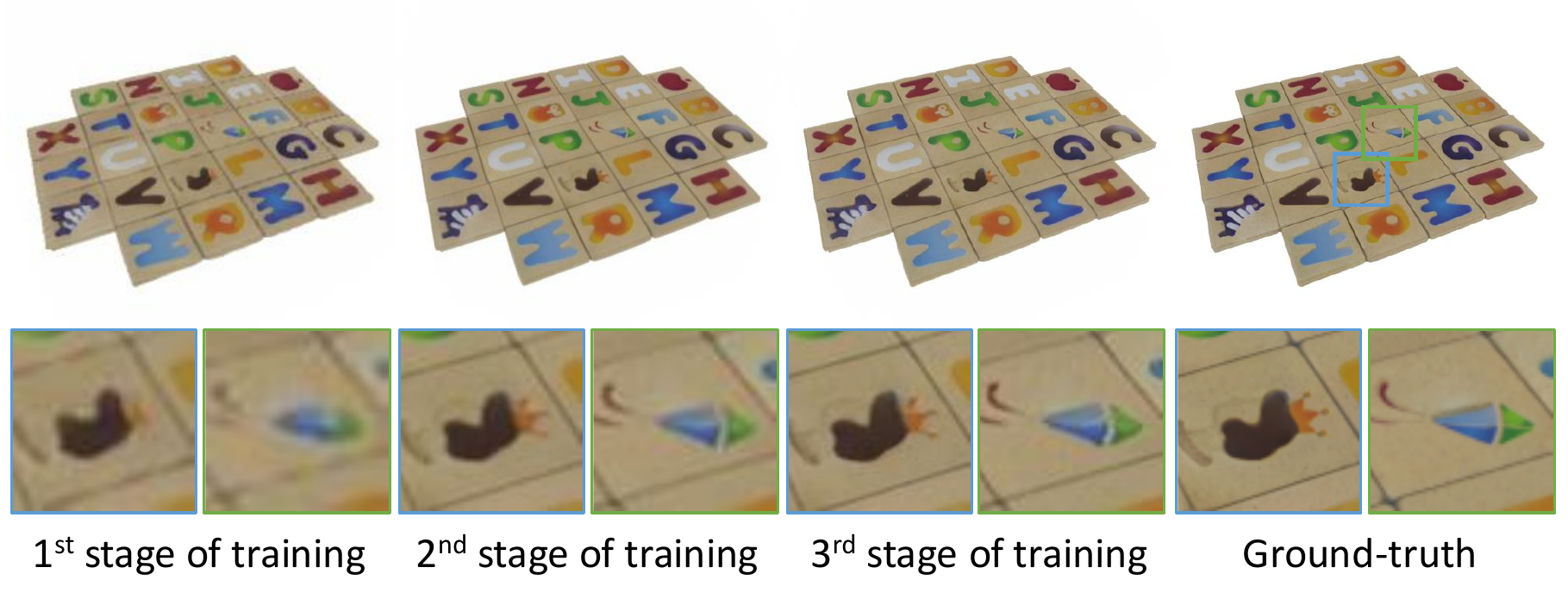}
\vspace{-0.2in}
\caption{Qualitative comparisons of view synthesis results after different stages of our coarse-to-fine training paradigm.}
\vspace{-0.1in}
\label{fig:qual_coarse_to_fine}
\end{figure}

\paragraph{Impacts of coarse-to-fine training} We test the network's reconstruction quality after different stages of training. We run the experiments on the GSO dataset rendered with uniform lighting. The quantitative results are summarized in Tab. \ref{tab:quant_coarse_to_fine}. We report the view synthesis metrics and chamfer distance after the $4^{\text{th}}$ update. The second stage of training significantly enhances texture details, while the second and third stages exhibit similar texture quality. However, the geometry quality in the third stage is better due to the incorporation of numerical normal loss. Fig. \ref{fig:qual_coarse_to_fine} visualizes view synthesis results from different stages of training.

\begin{table}
\caption{Quantitative comparisons of different stages of training for view synthesis under uniform lighting on \textbf{GSO} dataset}
    \label{tab:quant_coarse_to_fine}
    \vspace{-0.1in}
    \setlength{\tabcolsep}{4pt}
    \renewcommand{\arraystretch}{0.4} 
    \small
    \centering
    \begin{tabular}{lcccc}
    \toprule
     LIRM-hexa $4^{th}$ & PSNR ($\uparrow$) & SSIM ($\uparrow$) & LPIPS ($\downarrow$) & CD ($\downarrow$) \\
    \midrule
     Stage 1 & 27.56 & 0.924 & 0.113 & 0.120\\
     Stage 2 & \mrka{30.80} & \mrka{0.950} & 0.060 & 0.118 \\
     Stage 3 & 30.56 & 0.948 & \mrka{0.054} & \mrka{0.115} \\
     \bottomrule
    \end{tabular}
\end{table}

\begin{figure}
    \centering
    \includegraphics[width=\columnwidth]{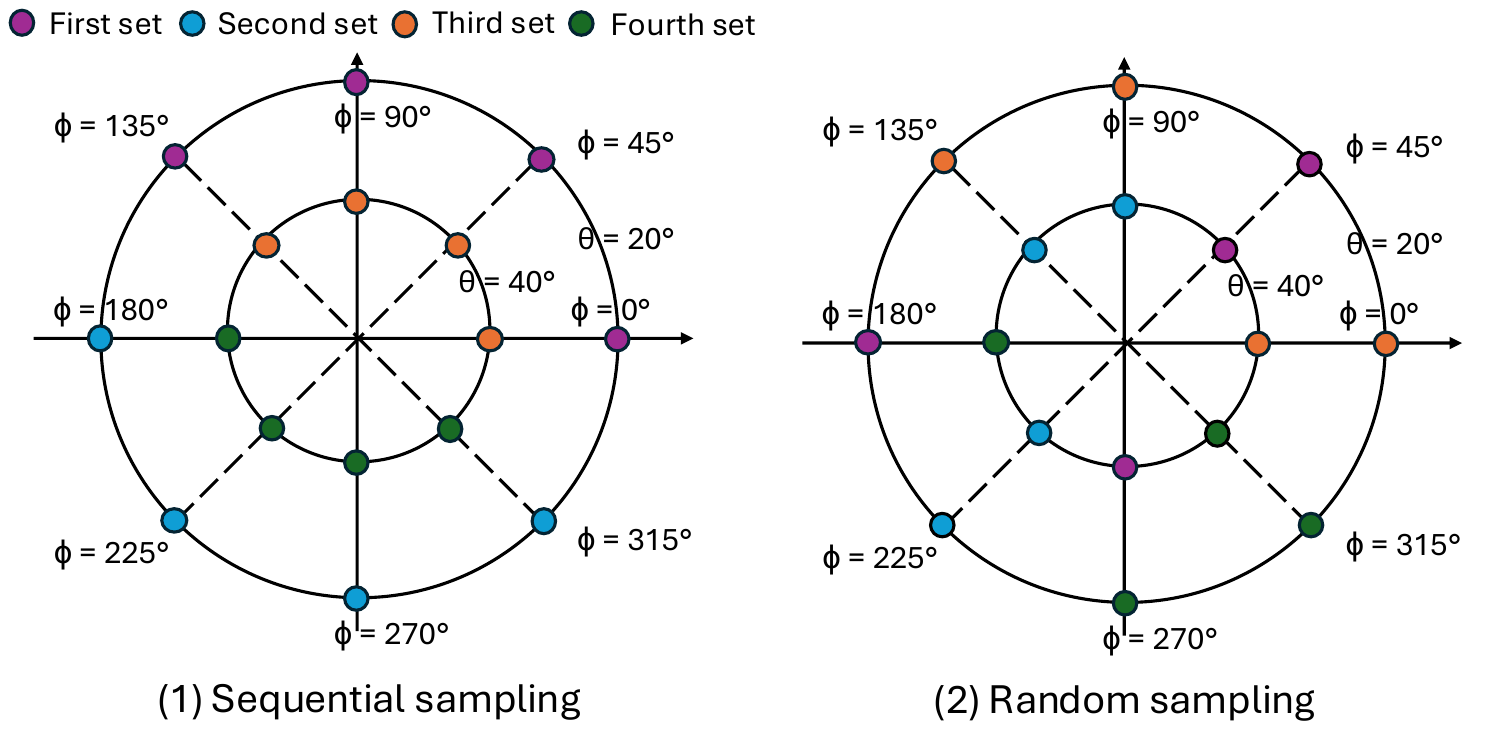}
    \vspace{-0.2in}
    \caption{Two different camera trajectories to test LIRM's update model. Random sampling is the default camera trajectory used in the main paper. $\theta$ and $\phi$ are elevation and azimuth angles.}
    \label{fig:cam_traj}
    \vspace{-0.1in}
\end{figure}

\begin{table}
    \caption{Quantitative comparisons of different camera trajectories for view synthesis under uniform lighting on \textbf{GSO} dataset. "Rd" and "Sq" represent random sampling and sequential sampling. The numbers of rows (1st to 4th) represent the number of updates performed by our model.}
    \label{tab:cam_traj_res}
    \vspace{-0.1in}
    \setlength{\tabcolsep}{4pt}
    \renewcommand{\arraystretch}{0.4} 
    \small
    \centering
    \begin{tabular}{lcccccc}
    \toprule
        & \multicolumn{2}{|c|}{PSNR ($\uparrow$)} & \multicolumn{2}{|c|}{SSIM ($\uparrow$)} & \multicolumn{2}{|c|}{LPIPS ($\downarrow$)} \\
        & Rd & Sq & Rd & Sq & Rd & Sq \\
     \midrule
     $1^{\text{st}}$  & 29.27 & 27.70  & 0.941 & 0.933 & 0.061 & 0.081 \\
     $2^{\text{nd}}$  & 30.48 & 30.09  & 0.947 & 0.946 & 0.056 & 0.060 \\
     $3^{\text{rd}}$  & \mrkb{30.65} & \mrka{30.66}  & \mrka{0.949} & \mrka{0.949} & \mrka{0.054} & \mrkb{0.055} \\
     $4^{\text{th}}$  & \mrkc{30.56} & \mrkc{30.56} & \mrkb{0.948} & \mrkb{0.948}  & \mrka{0.054} & \mrkb{0.055}  \\
     \bottomrule
    \end{tabular}
    \vspace{-0.1in}
\end{table}

\begin{figure}
    \centering
    \includegraphics[width=\columnwidth]{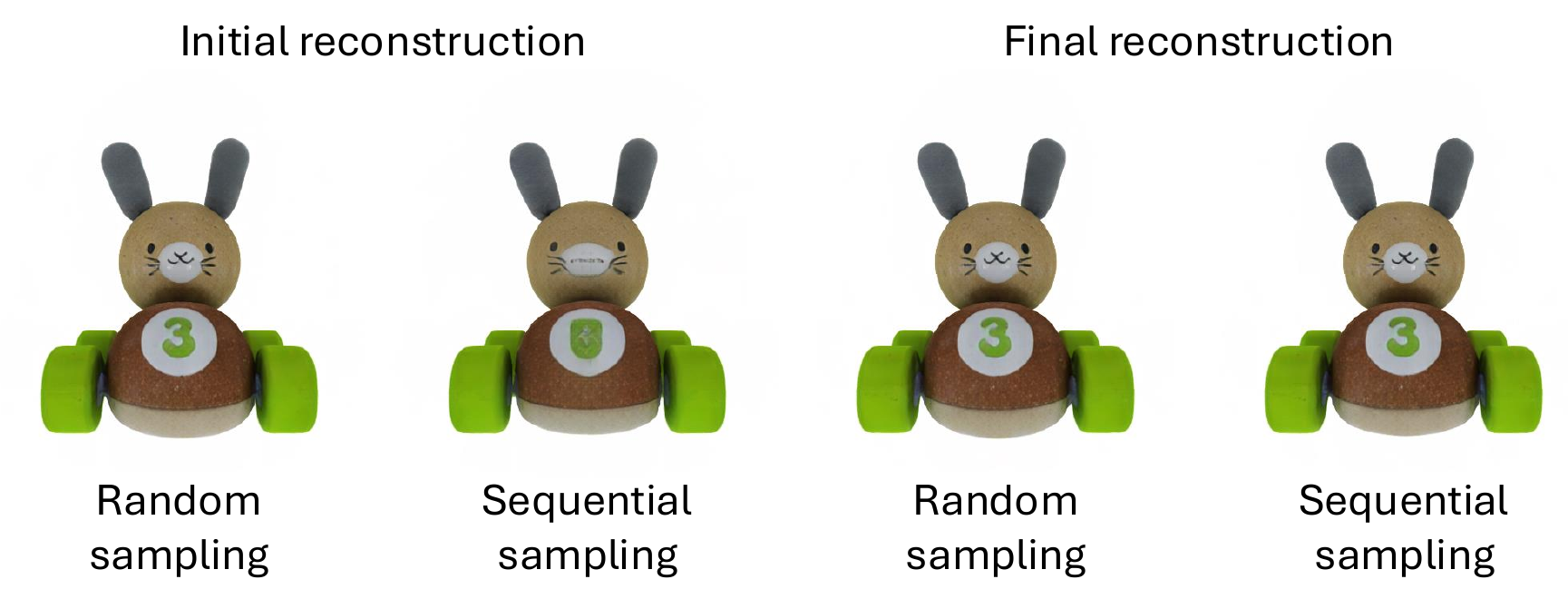}
    \vspace{-0.2in}
    \caption{Comparisons of reconstruction results under different camera trajectories. LIRM is robust to the order of input images and can converge to similar reconstruction results.}
    \vspace{-0.1in}
    \label{fig:cam_traj_res}
\end{figure}

\vspace{-0.1in}
\paragraph{Impacts of inference camera trajectories} For all the synthetic data experiments in the main paper, all the input views are fed to LIRM follow the same random order, which is shown in Fig. \ref{fig:cam_traj} (2). We test the impact of camera trajectory by feeding input images into LIRM sequentially, as shown in Fig. \ref{fig:cam_traj} (1). Qualitative and quantitative results of the new camera trajectory are summarized in Fig. \ref{fig:cam_traj_res} and Tab. \ref{tab:cam_traj_res} respectively. We observe that the initial reconstruction results are worse when we follow the new sequential order because those initial input images only observe one side of objects. However, our reconstruction errors converge to similar numbers after using all 16 images. It shows that LIRM update module is very robust to camera trajectories. 

\vspace{-0.2in}
\paragraph{Comparisons with a MeshLRM baseline} Prior state-of-the-art LRM-based mesh reconstruction method \cite{wei2024meshlrm} has not been open sourced yet. To compare with this strong baseline, we trained an LRM-VolSDF model with the same network architecture as \cite{wei2024meshlrm} using our newly created synthetic dataset built on Shutterstock \cite{Shutterstock}. Tab. \ref{tab:meshlrm_baseline} compares the baseline model with LIRM on GSO dataset rendered with uniform lighting. We observe that LIRM consistently performs better compared to the baseline model with different number of input images. Moreover, our update model enables us to utilize more input images without increasing GPU memory consumption, whereas the baseline model requires sending all images to the transformer simultaneously.

\begin{table}
    \caption{Quantitative comparisons for view synthesis under uniform lighting on \textbf{GSO} dataset with different number of images.}
    \label{tab:meshlrm_baseline}
    \vspace{-0.1in}
    \setlength{\tabcolsep}{6pt}
    \renewcommand{\arraystretch}{0.4} 
    \small
    \centering
    \begin{tabular}{lccc}
    \toprule
    4 images & PSNR ($\uparrow$) & SSIM ($\uparrow$) & LPIPS ($\downarrow$) \\
    \midrule
     Baseline & 28.72 & 0.940 & 0.070 \\
     LIRM-hexa $1^{\text{st}}$ & \mrka{29.27} & \mrka{0.941} & \mrka{0.061}  \\
     \midrule
      8 images & PSNR ($\uparrow$) & SSIM ($\uparrow$) & LPIPS ($\downarrow$) \\
      \midrule 
      Baseline & 30.19 & 0.947 & 0.061 \\
     LIRM-hexa $2^{\text{nd}}$  & \mrka{30.48} & \mrka{0.947} & \mrka{0.056} \\
        \midrule
      12 images & PSNR ($\uparrow$) & SSIM ($\uparrow$) & LPIPS ($\downarrow$) \\
      \midrule 
      Baseline & 30.50 & 0.948 & 0.059 \\
     LIRM-hexa $3^{\text{rd}}$  & \mrka{30.65} & \mrka{0.949} & \mrka{0.054} \\      
     \bottomrule
    \end{tabular}
    \vspace{-0.1in}
\end{table}

\begin{table}
\caption{
Benchmarks for BRDF parameters. We aligned albedo scales before evaluation, following Stanford-ORB.
}
\label{tab:optim_brdf}
\centering
\vspace{-0.1in}
\setlength{\tabcolsep}{6pt}
\renewcommand{\arraystretch}{0.4} 
\small
\centering
\begin{tabular}{l|ccc|c}
\toprule
 & \multicolumn{3}{|c|}{PSNR ($\uparrow$)} & LPIPS ($\downarrow$) \\
 \midrule
  & $\*a$ & $\*r$ & $\*m$ & $\*a$ \\ 
 \midrule
 LIRM-diff 3$^\text{rd}$ & \mrka{28.20} & \mrka{27.10} & \mrka{22.17} & 0.086\\
 \midrule
 InvRender & 25.14 & 16.15 & - & 0.122\\  
  \midrule
 NVDiffrecMc & 25.58 & 17.11 & 15.00 & 0.114 \\
  \midrule
 Neural-PBIR & 24.62 & 13.63 & - & \mrka{0.071}\\
 \bottomrule 
\end{tabular}
\end{table}

\begin{figure}
    \centering
    \includegraphics[width=\columnwidth]{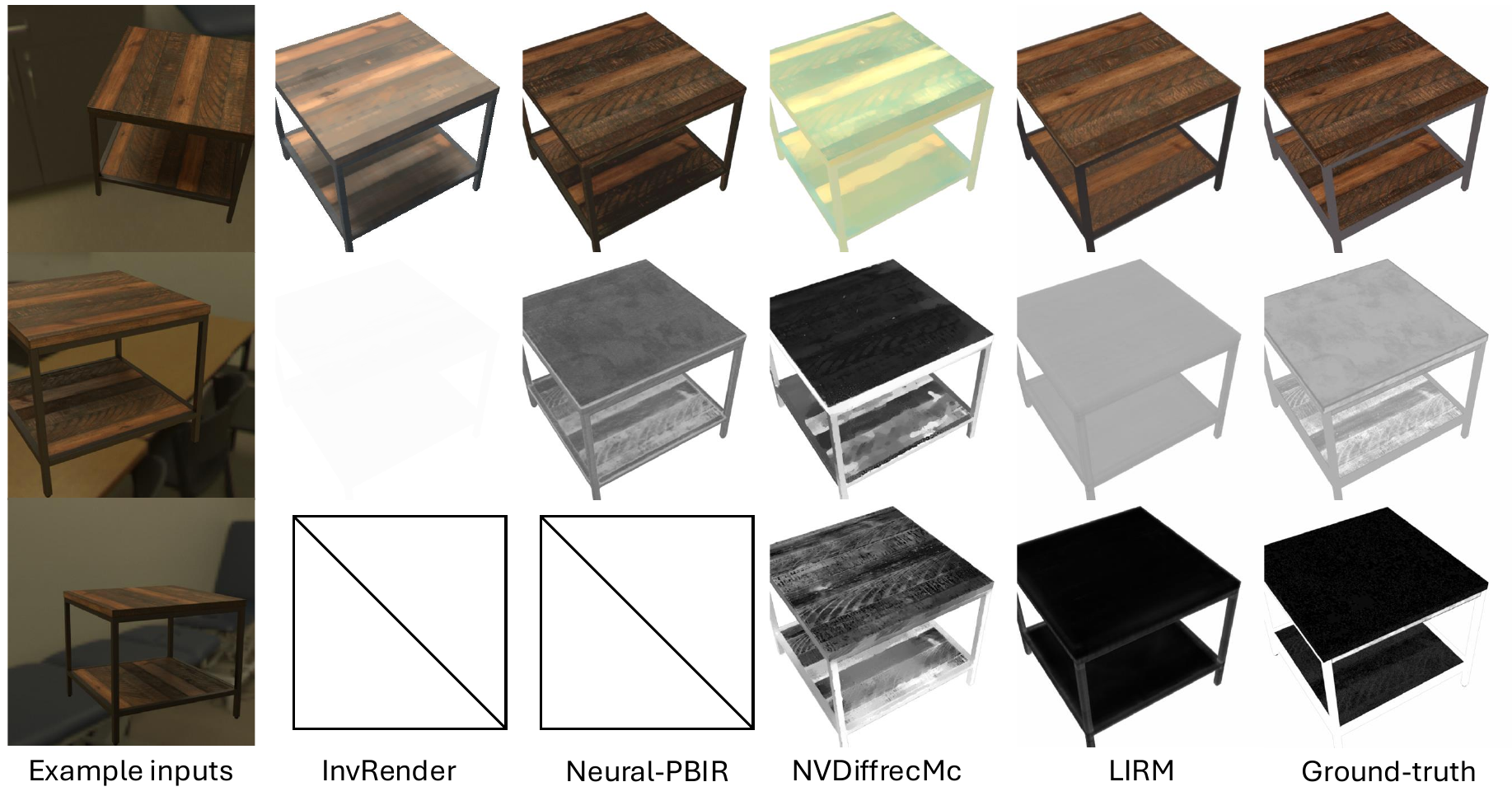}
    \vspace{-0.3in}
    \caption{BRDF prediction results. LIRM better recovers BRDF of the bottom board of the table, which is under a shadow in inputs.}
    \vspace{-0.1in}
    \label{fig:optim_brdf}
\end{figure}

\vspace{-0.2in}
\paragraph{Comparisons with optimization-based inverse rendering methods} As optimization-based inverse rendering methods usually take several minutes or even hours to process one model, it is unrealistic to add them in quantitative comparisons in Tab. \ref{tab:quant_env_abo} and Tab. \ref{tab:quant_env_dtc}, where we test 1000 models. Instead, we evaluate BRDF reconstruction on a much smaller synthetic data by randomly selecting 10 models from ABO \cite{collins2022abo} and DTC \cite{Dong_2025_CVPR} datasets (5 each). For each model, we render 100 images for optimization and 10 for testing. We randomly select 12 images as inputs to LIRM. Quantitative and qualitative results are summarized in Tab. \ref{tab:optim_brdf} and Fig. \ref{fig:optim_brdf} respectively. We observe that LIRM can reconstruct spatially varying BRDF parameters accurately, especially working well in decomposing lighting and shadows from BRDF, similar to what we observe in Fig. \ref{fig:exp_brdf_with_gt}. Note that when reporting quantitative numbers on relighting and diffuse albedo, we follow \cite{kuang2024stanford} to compute separate scales for RGB channels. Therefore, the advantages of better lighting decomposition of LIRM have not been properly revealed in the above numbers yet.

\end{document}